\documentclass{article} % For LaTeX2e
\usepackage{iclr2026_conference,times}
\iclrfinalcopy
\usepackage{amsmath,amsfonts,bm}

\def\eqref#1{equation~\ref{#1}}
\def\1{\bm{1}}

\DeclareMathAlphabet{\mathsfit}{\encodingdefault}{\sfdefault}{m}{sl}
\SetMathAlphabet{\mathsfit}{bold}{\encodingdefault}{\sfdefault}{bx}{n}

\usepackage{hyperref}
\usepackage{wrapfig}
\usepackage{url}
\usepackage{graphicx}
\usepackage{booktabs}
\usepackage{multirow}
\usepackage{graphicx}
\usepackage[dvipsnames]{xcolor}
\usepackage[table]{xcolor}
\usepackage{amssymb}
\usepackage{seqsplit}
\usepackage{listings}
\usepackage{xcolor}
\usepackage{colortbl}
\usepackage{subcaption}

\usepackage{pgfplots}
\pgfplotsset{compat=1.18}

\title{PALADIN: Self-Correcting Language Model Agents to Cure Tool-Failure Cases}

\author{
  Sri Vatsa Vuddanti \\
  \texttt{srivatsa644@gmail.com}
  \And
  Aarav Shah \\
  \texttt{magnifed1234@gmail.com}
  \And
  Satwik Kumar Chittiprolu \\
  \texttt{satchi427@gmail.com}
  \And
  Tony Song \\
  \texttt{hexatonysong@gmail.com}  
  \And
  Sunishchal Dev \\
  \texttt{dev@algoverseairesearch.org}
    \And
  Kevin Zhu \\
  \texttt{kevin@algoverseacademy.com}    
  \And
  Maheep Chaudhary\thanks{Project Lead} \\
  \texttt{maheepchaudhary.research@gmail.com}
}

\begin{document}

\maketitle

\begin{abstract}
Tool-augmented language agents frequently fail in real-world deployment due to tool malfunctions—timeouts, API exceptions, or inconsistent outputs—triggering cascading reasoning errors and task abandonment. Existing agent training pipelines optimize only for success trajectories, failing to expose models to the tool failures that dominate real-world usage.
We propose \textbf{PALADIN}, a generalizable framework for equipping language agents with robust failure recovery capabilities. PALADIN trains on 50,000+ recovery-annotated trajectories constructed via systematic failure injection and expert demonstrations on an enhanced ToolBench dataset. Training uses LoRA-based fine-tuning to retain base capabilities while injecting recovery competence. At inference, PALADIN detects execution-time errors and retrieves the most similar case from a curated bank of 55+ failure exemplars aligned with ToolScan's taxonomy, then executes the corresponding recovery action.
This approach generalizes to novel failures beyond the training distribution, retaining 95.2\% recovery performance on unseen tool APIs. Evaluation across PaladinEval and ToolReflectEval demonstrates consistent improvements in Recovery Rate (RR), Task Success Rate (TSR), Catastrophic Success Rate (CSR), and Efficiency Score (ES). PALADIN improves RR from 32.76\% to 89.68\% (+57\% relative) over ToolBench and outperforms the strongest baseline CRITIC (76.34\%) by +13.3\%. Against vanilla agents, PALADIN achieves 89.86\% RR (+66\% relative improvement from 23.75\%). These results establish PALADIN as an effective method for building fault-tolerant agents capable of robust recovery in real-world tool environments.
\end{abstract}

% \noindent\textbf{Keywords:} agent robustness, error recovery, large language models, real-world deployment, benchmark, tool use, reinforcement learning, supervised fine-tuning, resilience metrics \\

\section{Introduction}

Tool-augmented LLM agents are increasingly deployed in real-world environments for automation, reasoning, and decision-making. While benchmarks like ToolBench~\citep{qin2023toolllm} and Gorilla~\citep{patil2023gorilla} demonstrate strong performance in clean settings, agents often fail under realistic conditions—APIs time out, tool outputs are malformed, and calls silently fail. Despite occasional inclusion of noisy inputs or low-confidence reasoning, current training pipelines do not explicitly expose agents to structured tool failures. This results in brittle behavior: agents silently hallucinate success or deadlock when tool calls fail, contradicting real-world operational requirements.

% We identify \textbf{execution-level robustness}—the ability to detect, diagnose, and recover from runtime failures—as a central unsolved challenge for tool-augmented agents.
We identify execution-level robustness—the ability to detect, diagnose, and recover from runtime failures—as a central unsolved challenge for tool-augmented agents.
Existing work focuses on \textit{comprehension robustness}, improving plans or tool formats, but leaves agents brittle when tools malfunction, leading to deadlocks, loops, or hallucinated success. Approaches such as ToolReflect~\citep{polyakov-etal-2025-toolreflection} and CRITIC~\citep{gao2023secrets} introduce reflective corrections, but these are reactive and limited to call-level fixes. 
% In contrast, \textbf{PALADIN} redefines the training signal with recovery-annotated trajectories and augments inference with taxonomy-guided recovery, enabling proactive and generalizable error handling.
Furthermore, evaluation metrics traditionally focus on task success without measuring recovery capabilities, limiting our understanding of agent robustness. 
We argue that execution-level robustness must be explicitly taught through systematic training
To address this, we propose PALADIN, a generalizable method for teaching LLM agents to recover from tool-use failures via trajectory-level supervision. 
% PALADIN uses automated trace analysis, targeted failure injection, and recovery-annotated training to instill resilient behavior. Furthermore, PALADIN detects execution-time errors and links them to their nearest counter part in a set of 55+ failure exemplars. 
\begin{wrapfigure}{r}{0.6\textwidth}
    \centering
    \includegraphics[width=0.58\textwidth]{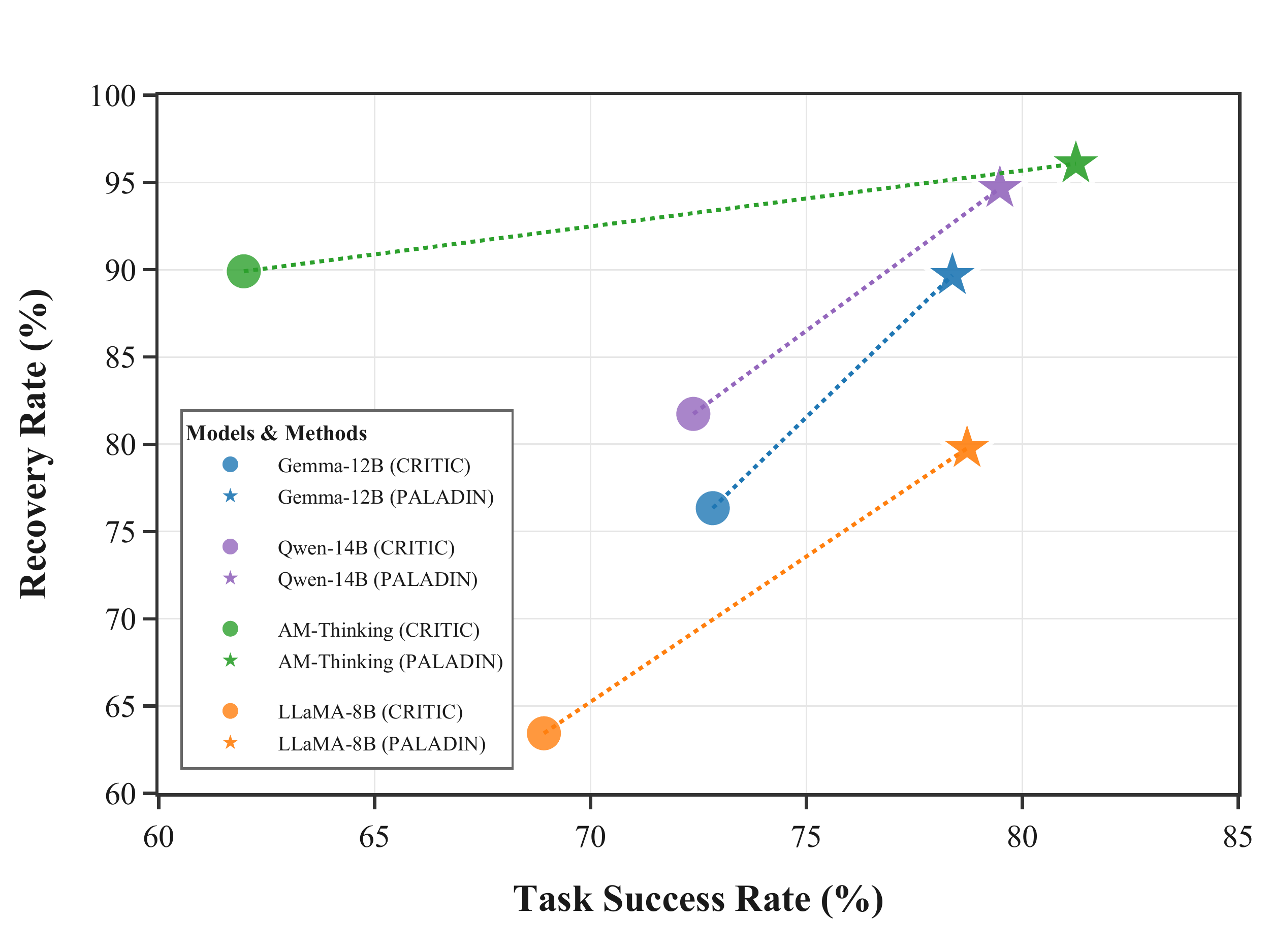}
    \caption{Recovery Rate vs Task Success Rate comparison between CRITIC and PALADIN across different LLMs. Stars indicate PALADIN results, circles indicate CRITIC baseline. Dotted lines show improvement trajectories from CRITIC to PALADIN for each model. PALADIN consistently achieves higher task success rates while maintaining superior error recovery capabilities.}
    \label{fig:paladin_results}
\end{wrapfigure}

PALADIN combines systematic failure injection with recovery-annotated training, using GPT-5 API to simulate multi-turn tool use and generate recovery annotations, resulting in 50,000+ trajectories spanning diverse error modes from ToolScan's taxonomy~\citep{kokane2025toolscanbenchmarkcharacterizingerrors}. At inference, PALADIN detects execution errors and retrieves recovery strategies from 55+ curated failure exemplars.
We also define four metrics to evaluate agent robustness: \textbf{Task Success Rate (TSR)}, \textbf{Recovery Rate (RR)}, \textbf{Catastrophe Success Rate (CSR)}, and \textbf{Efficiency Score (ES)}, with the latter three being novel proposals.

In evaluations with deterministic error injection, conducted using GPT-5 API as a tool-use simulator, PALADIN improves RR and CSR, at the cost of a lower ES compared to other methods.
model sizes,demonstrating that execution-level robustness is a learnable and scalable behavior, and positioning PALADIN as a significant step towards design of resilient LLM agents.

\textbf{Key Contributions:}

\begin{itemize}

\item \textbf{PALADIN}: A training framework combining systematic failure injection with taxonomy-guided recovery, enabling agents to learn generalizable error handling from 50,000+ failure-recovery trajectories.

\item \textbf{Recovery-focused evaluation}: Three novel metrics, including \textit{Recovery Rate, Catastrophe Success Rate}, and \textit{Efficiency Score}; and  \textit{PaladinEval} benchmark for systematic robustness assessment.

\item \textbf{Inference-time recovery}: A retrieval-based approach that matches runtime tool failures to similar recovery examples, maintaining 95.2\% of training performance when generalizing to new APIs.

\item \textbf{Empirical validation}: Demonstrating execution-level robustness as learnable, with 57-66\% relative improvements in recovery rates across model sizes.

\end{itemize}

\section{Methodology}

\begin{figure}[t]
    \includegraphics[width=\linewidth]{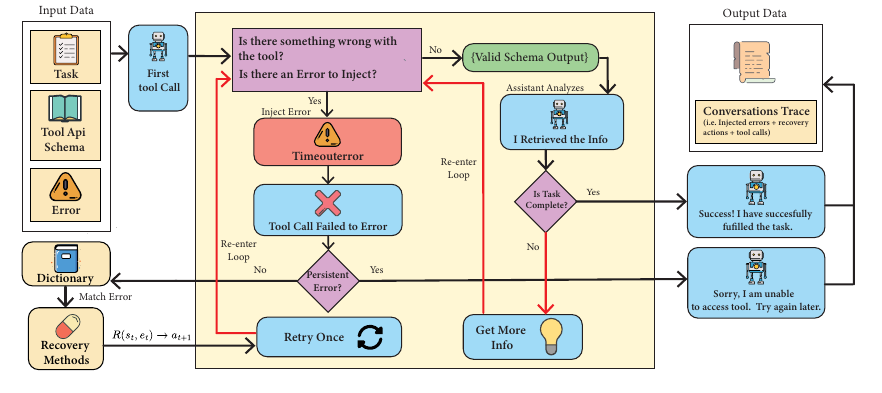}
    \caption{Our tool-use simulator with integrated error injection and recovery mechanisms. (a) depicts the static architecture, where errors are injected into tool calls and handled via a recovery dictionary. (b) details the dynamic execution loop, capturing assistant reasoning, recovery actions, and final outcomes. This design allows controlled, reproducible evaluation of LLM resilience to tool failures.}
    \label{fig:simulator}
\end{figure}

PALADIN is a method for equipping tool-augmented LLM agents with execution-level robustness\textemdash the ability to detect, diagnose, and recover from runtime failures. The method consists of (1) formalizing tool failures, (2) constructing recovery-annotated training data via failure injection, (3) fine-tuning with a recovery-aware objective, and (4) augmenting inference with taxonomy-driven retrieval. Figure~\ref{fig:simulator} presents a high-level overview.

\subsection{Problem Setup}
We model a tool-augmented agent as a policy $\pi_\theta$ producing trajectories $\tau = 
[(s_1,a_1),\dots,(s_T,a_T)]$, where $s_t$ are states and $a_t$ may include tool calls. Failures $f \in \mathcal{F}$ occur when tool execution yields timeouts, malformed outputs, API errors, or inconsistencies. Existing training ignores such events, yielding brittle policies. Our goal is to learn $\pi^*$ that maximizes task success while remaining robust to failures:
\[
\pi^* = \arg\max_\pi \; \mathbb{E}_{\tau \sim \pi} [ \text{TSR}(\tau) - \alpha \cdot \text{CSR}(\tau) ].
\]

\subsection{Failure Injection and Recovery Annotation}

%\mch{[\textit{written it with claude, pleas review, make the below para clear with it}] To construct PALADIN's dataset, we systematically augmented ToolBench trajectories by injecting failures corresponding to ToolScan's seven canonical error classes. Each trajectory τ is decomposed into task description T, available toolset A, execution trace C, and error signal E (if present). When E ≠ ∅, we truncate C at the failure point and apply GPT-4-guided recovery annotation:
%$$f_{repair}(T, A, C, E) → C'$$
%where C' represents the corrected trajectory with explicit recovery actions. The teacher model is prompted with failure context and generates appropriate recovery strategies (retry with modified parameters, tool substitution, or graceful degradation).}

To construct PALADIN’s dataset, we systematically augmented ToolBench trajectories by injecting failures aligned to the ToolScan~\citep{kokane2025toolscanbenchmarkcharacterizingerrors} taxonomy of seven canonical error classes. Each trajectory is parsed into task $T$, toolset $\mathcal{A}$, trace $C$, and error signal $E$. If $E \ne \varnothing$, we truncate $C$ at the failure point and pass it to a GPT-5 Teacher (e.g. GPT-5 API prompted to insert proper recovery actions), which applies error-guided rewriting with the goal of creating realistic error trajectories.
\[
f_{\text{repair}}(T, \mathcal{A}, C, E) \rightarrow C'.
\]
Otherwise, $C$ is finalized cleanly:
\[
f_{\text{finalize}}(T, \mathcal{A}, C) \rightarrow C'.
\]

All output traces ($C'$) are stored alongside their error signal ($E$) and recovery metadata, yielding over 50K recovery-annotated trajectories. (see Appendix~\ref{app:traces} for detailed augmentation pipeline)
Figure~\ref{fig:failure_injection} illustrates this pipeline.

We also curated a recovery dictionary, a collection of 55+ exemplar failures and associated recovery strategies. These exemplars operationalize the ToolScan taxonomy into practical, reusable supervision signals.

\subsection{PALADIN Training Objective}

PALADIN fine-tunes base LLMs using a causal language modeling objective augmented with recovery supervision:

\[
\mathcal{L}_{\text{PALADIN}} = \mathcal{L}_{\text{SFT}} + \lambda \, \mathcal{L}_{\text{rec}},
\]

where $\mathcal{L}_{\text{SFT}}$ is the standard negative log-likelihood over successful trajectories, and $\mathcal{L}_{\text{rec}}$ is the same objective restricted to tokens following \texttt{Recovery:} tags that mark corrective steps. Training sequences are serialized in ToolBench format, and LoRA \citep{hu2021lora} adapters provide parameter-efficient adaptation while preserving base competence.

\subsection{Inference-Time Recovery via Taxonomic Retrieval}
During execution, PALADIN detects runtime errors and links them to a curated exemplar bank $ \mathcal{E} = \{(f_i, r_i)\}_{i=1}^{55}$, consisting of 55+ exemplar failures $f_i$ with different recovery action $r_i$ given in Appendix~\ref{app:runtime_e} and Appendix~\ref{app:python_dic}. Given an observed failure $f_{obs}$, PALADIN retrieves the most similar exemplar:
\[
f_{ref} = \arg\min_{f_i \in \mathcal{E}} d(f_{obs}, f_i),
\]
where $d$ is a similarity metric over error signatures. The corresponding recovery action $r_i$ is executed, guiding the agent back to a stable trajectory. This enables proactive recovery across diverse failure scenarios (see how the recovery dictionary was developed at Appendix~\ref{app:dataset}). 

\subsection{Recovery as Implicit Policy Learning}
Although PALADIN is trained end-to-end, its behavior (see Appendix~\ref{app:expanded} for a better overview on PALADIN's thought process) can be viewed as an implicit recovery policy:
\[
R(s_t, e_t) \rightarrow a_{t+1},
\]
where $s_t$ is the dialogue context, $e_t$ is the observed failure, and $a_{t+1}$ is the recovery action (retry, reformat, switch tools, or terminate). Unlike reflective methods such as ToolReflect or call-level critique approaches such as CRITIC, which introduce reactive adjustments at inference time, PALADIN learns recovery behaviors directly from annotated trajectories. This yields proactive, compositional recovery strategies (e.g., retries $\rightarrow$ tool switch $\rightarrow$ graceful termination) that generalize across failure types.

\subsection{Metrics for Execution Robustness}

We evaluate PALADIN with four metrics:

\begin{equation*}
\begin{aligned}
\text{TSR} &= \frac{\# \text{successful tasks}}{\text{total tasks}}, \\
\text{RR} &= \frac{\text{\# failures recovered}}{\text{failures encountered}}, \\
\text{CSR} &= 1 - \frac{\text{\# hallucinated successes}}{\text{\# total failures}}, \\
\text{ES} &= \frac{1}{\text{average \# steps to complete task}}.
\end{aligned}
\end{equation*}

% Equivalently, we normalize our data with the Catastrophe Success Rate (CSR) as:
% \[
% \text{CSR} = 1 - \text{FCR},
% \]
% which represents the fraction of task failures that were correctly declared as failures.  

RR, CSR, and ES are novel contributions introduced to capture execution-level robustness beyond traditional task success.

\section{Experiment and Evaluation Setup}

\begin{figure}[h]
\begin{center}
\includegraphics[width=1\linewidth]{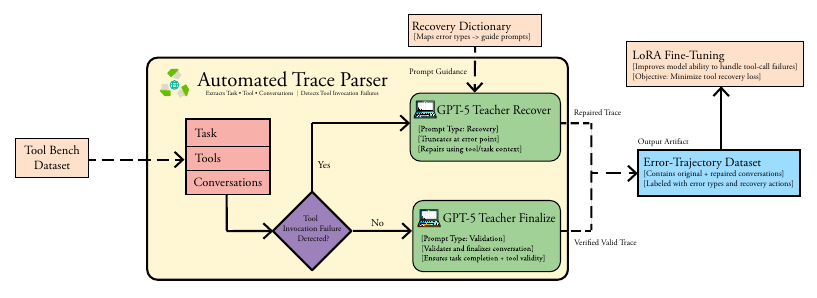}
\end{center}
\caption{Trace repair pipeline for constructing the Error-Trajectory Dataset. Each ToolBench trace is truncated at the first failure, then repaired or finalized via GPT-guided recovery. Outputs are stored with recovery metadata to construct PALADIN’s training corpus.}
\label{fig:failure_injection}
\end{figure}

\subsection{Benchmarks and Evaluation Protocol}

We evaluate PALADIN on three complementary benchmarks, designed to capture both controlled and naturalistic failure scenarios.

\paragraph{PaladinEval.} 
PaladinEval (benchmark design in Appendix~\ref{app:paladineval})
is derived from ToolBench’s evaluation dataset, and explicit failure injections from our own dataset. Each evaluation instance is constructed by extracting the original prompt from ToolBench and pairing it with a runtime error assigned by a teacher GPT model (see Appendix~\ref{app:dataset}). The resulting dataset contains a mix of clean tool-use traces and failure-augmented traces, enabling measurement of both normal execution and recovery. To support consistent grading, we provide evaluation guidelines specifying expected recovery strategies, which condition an automated grader (GPT-5 API) to score outputs along our four robustness metrics.

\paragraph{Generalization Set.}
To assess out-of-distribution robustness, we additionally construct a generalization dataset by sampling unused ToolBench~\citep{qin2023toolllm} evaluation prompts and injecting novel errors not present in the curated 55+ exemplar recovery dictionary. This ensures PALADIN is tested on recovery behaviors that were unseen during training.

\paragraph{ToolReflectEval.}
We further evaluate on the ToolReflect benchmark~\citep{polyakov-etal-2025-toolreflection}, which introduces reflective corrections for tool-use errors. To ensure comparability, we adopt their protocol of allowing up to three attempts per tool call. We directly incorporate all ToolAlpaca-derived prompts from their evaluation dataset without modification or cherry-picking. The only adaptation is re-execution within our simulator, which enables generation of full conversational traces and consistent grading under our robustness metrics. This ensures our evaluation remains faithful to ToolReflect’s task setup while allowing direct comparison against PALADIN.

\subsection{Simulation Environment}

To evaluate PALADIN under realistic tool-use conditions, we design a simulation environment that faithfully executes agent–tool interactions and introduces controlled failures. The environment reproduces the standard ToolBench execution pipeline while augmenting it with failure injection, recovery logging, and automated grading implementation details in Appendix~\ref{app:implementation}. Figure~\ref{fig:simulator} illustrates the workflow.

Each evaluation begins with a prompt $P$ and toolset $\mathcal{A}$. The environment executes tool calls sequentially, returning either successful outputs or injected failures $f \in \mathcal{F}$. Failures are generated either (i) systematically, via the PaladinEval injection protocol, or (ii) naturally, by replaying error-prone ToolReflectEval prompts. At each step, the environment logs the dialogue context $C_t$, the observed error $e_t$, and any recovery attempt $a_t$.

To ensure consistent scoring, we provide simulator outputs to an automated grader powered by GPT-5 API, conditioned on explicit evaluation guidelines to calculate different metrics. These guidelines specify valid recovery strategies, prohibited behaviors (e.g., hallucinated success), and metric definitions, enabling the grader to compute Task Success Rate (TSR), Recovery Rate (RR), Catastrophic Success Rate (CSR), and Efficiency Score (ES). 
This environment provides three advantages: (1) easy replication of tool-call behavior with both clean and failure-augmented traces, (2) Everything is graded on the same terms via guideline-conditioned automated evaluation, and (3) flexibility in allowing integration of external datasets such as ToolReflectEval without modifying their task structure.

\subsection{Baseline Agents}

We compare PALADIN against four baselines spanning different supervision paradigms and recovery strategies, with 
vanilla agent being raw pretrained language model
as baseline with other method like CRITIC \cite{gao2023secrets}, ToolBench Agent \cite{qin2023toolllm}, and ToolReflect Agent \cite{polyakov-etal-2025-toolreflection}.  
CRITIC (Appendix~\ref{app:critic}) accesses recovery protocols with probability $p=0.7$, approximating its partial ability to retrieve external signals while remaining reactive.
ToolBench Agent
replicates the ToolBench agent trained on the full ToolBench dataset of successful trajectories, being effective on clean traces but unable to recover from failures.
On the other hand ToolReflect Agent adapated from ToolReflect, is trained on error–correction pairs to label tool calls as valid/invalid. It performs iterative self-correction but remains limited to local, reactive adjustments.
These baselines collectively probe PALADIN’s contributions across supervision (clean vs. recovery-annotated), recovery strategy (reactive vs. proactive), and model integration (training-only vs. inference-time retrieval).

% \subsection{Evaluation Metrics}
% We report four metrics that together capture task effectiveness, robustness, safety, and efficiency, namely: Task Success Rate (TSR), Recovery Rate (RR), Catastrophe Success Rate (CSR), and Efficiency Score (ES).
% \begin{itemize}
%     \item \textbf{Task Success Rate (TSR)} — fraction of tasks completed successfully.
%     \item \textbf{Recovery Rate (RR)} — fraction of failures that were successfully recovered.
%     \item \textbf{Catastrophe Success Rate (CSR)} — the percentage of failures in which the model does not hallucinate.
%     \item \textbf{Efficiency Score (ES)} — inverse of average steps per task, rewarding concise recoveries.
% \end{itemize}
% \mch{Please define Catastrophe Success Rate (1-Failure Catastrophe Rate)}

\section{Results}
\begin{table*}[httb]
\caption{Main results across models and datasets. All metrics (Recovery, Task Success, CSR, Efficiency) are normalized such that higher is better. PALADIN achieves consistently strong safety and recovery performance, with modest efficiency tradeoffs}
\label{tab:results}
\begin{center}
\begin{scriptsize}
% \begin{sc}
% \begin{adjustbox}{width=1\linewidth} 
\begin{tabular}{c|c|c|c|c|c|c|c}

\toprule

\multirow{2.5}{*}{\textbf{{\shortstack{Pretrained\\LLM}}}} & 
\multirow{2.5}{*}{\textbf{{\shortstack{Evaluation \\ Datasets}}}} & 
\multirow{2.5}{*}{\textbf{{\shortstack{Evaluation \\ Metrics}}}} & 
\multicolumn{5}{c}{\textbf{Scores $(\uparrow)$}} \\

\cmidrule(lr){4-8}

& & & Vanilla & CRITIC & ToolReflect & ToolBench &  \cellcolor{gray!10}Paladin (Ours)\\

\midrule
    
\multirow{8}{*}{\centering \shortstack{Gemma-3\\12B-Instruct}} & \multirow{4}{*}{\shortstack{Paladin \\ Eval}}& Recovery Rate & 23.75\%  & 76.34\% & 65.86\% & 32.76\% & \cellcolor{gray!10}\textbf{89.68\%} {\tiny \textcolor{blue}{+13.34\%}} \\

& & Task Success Rate & 23.62\% & 72.83\% & 61.42\% & 57.4\% & \cellcolor{gray!10}\textbf{78.38\%} {\tiny \textcolor{blue}{+5.55\%}} \\

& & Catastrophic Success Rate & 29.00\% & 73.30\% & 70.27\% & 68.37\% & \cellcolor{gray!10}\textbf{82.55\%} {\tiny \textcolor{blue}{+9.25\%}} \\s

& & Efficiency Score & \textbf{0.476} & 0.348 & 0.288 & 0.221 & \cellcolor{gray!10}0.312 {\tiny \textcolor{blue}{-34.45\%}} \\

\cmidrule(lr){2-8}

& \multirow{4}{*}{\shortstack{ToolReflect\\Eval}} & Recovery Rate & 22.80\% & 73.29\% & 63.23\% & 31.45\% & \cellcolor{gray!10}\textbf{86.09\%} {\tiny \textcolor{blue}{+12.80\%}} \\

& & Task Success Rate & 22.56\% & 69.55\% & 58.66\% & 54.82\% & \cellcolor{gray!10}\textbf{83.45\%} {\tiny \textcolor{blue}{+13.90\%}} \\

& & Catastrophic Success Rate & 26.16\% & 72.23\% & 69.08\% & 67.10\% & \cellcolor{gray!10}\textbf{81.85\%} {\tiny \textcolor{blue}{+9.62\%}} \\

& & Efficiency Score & \textbf{0.508} & 0.370 & 0.307 & 0.235 & \cellcolor{gray!10}0.332 {\tiny \textcolor{blue}{-34.65\%}} \\

\midrule

\multirow{8}{*}{\shortstack{Qwen-2.5-\\14B-Instruct}} & \multirow{4}{*}{\shortstack{Paladin \\ Eval}} & Recovery Rate & 37.68\% & 81.74\% & 73.66\% & 33.48\% & \cellcolor{gray!10}\textbf{94.67\%} {\tiny \textcolor{blue}{+12.93\%}} \\

& & Task Success Rate & 37.53\% & 72.38\% & 74.36\%  & 60.41\% & \cellcolor{gray!10}\textbf{79.48\%} {\tiny \textcolor{blue}{+5.12\%}} \\

& & Catastrophic Success Rate & 56.53\% & 87.14\% & 83.85\% & 67.88\% & \cellcolor{gray!10}\textbf{94.57\%} {\tiny \textcolor{blue}{+7.43\%}} \\

& & Efficiency Score & 0.313 & 0.329 & 0.312 & 0.339 & \cellcolor{gray!10}\textbf{0.351} {\tiny \textcolor{blue}{+3.54\%}} \\

\cmidrule(lr){2-8}

& \multirow{4}{*}{\shortstack{ToolReflect\\Eval }} & Recovery Rate & 36.17\% & 78.47\% & 70.71\% & 32.14\% & \cellcolor{gray!10}\textbf{92.88\%} {\tiny \textcolor{blue}{+14.41\%}} \\

& & Task Success Rate & 35.85\% & 69.13\% & 71.02\% & 57.70\% & \cellcolor{gray!10}\textbf{75.19\%} {\tiny \textcolor{blue}{+4.17\%}} \\

& & Catastrophic Success Rate & 54.88\% & 81.63\% & 83.20\% & 66.60\% & \cellcolor{gray!10}\textbf{94.35\%} {\tiny \textcolor{blue}{+11.15\%}} \\

& & Efficiency Score & 0.334 & 0.350 & 0.333 & 0.361 & \cellcolor{gray!10}\textbf{0.375} {\tiny \textcolor{blue}{+3.73\%}} \\

\midrule

\multirow{8}{*}{\shortstack{AM-Thinking\\V1}} & \multirow{4}{*}{\shortstack{Paladin \\ Eval}} & Recovery Rate & 49.87\% & 89.91\% & 87.23\% & 51.37\% & \cellcolor{gray!10}\textbf{96.08\%} {\tiny \textcolor{blue}{+6.17\%}} \\

& & Task Success Rate & 52.93\% & 61.97\% & 72.88\% & 56.83\% & \cellcolor{gray!10}\textbf{81.24\%} {\tiny \textcolor{blue}{+8.36\%}} \\

& & Catastrophic Success Rate & 60.24\% & 81.33\% & 71.32\% & 80.84\% & \cellcolor{gray!10}\textbf{88.65\%} {\tiny \textcolor{blue}{+7.32\%}} \\

& & Efficiency Score & 0.415 & \textbf{0.420} & 0.297 & 0.319 & \cellcolor{gray!10}0.380 {\tiny \textcolor{blue}{-9.52\%}} \\

\cmidrule(lr){2-8}

& \multirow{4}{*}{\shortstack{ToolReflect\\Eval}} & Recovery Rate & 47.88\% & 86.31\% & 83.74\% & 49.32\% & \cellcolor{gray!10}\textbf{92.24\%} {\tiny \textcolor{blue}{+5.93\%}} \\

& & Task Success Rate & 50.55\% & 78.64\% & 79.41\% & 65.25\% & \cellcolor{gray!10}\textbf{77.98\%} {\tiny \textcolor{blue}{-1.43\%}} \\

& & Catastrophic Success Rate & 62.65\% & 67.42\% & 65.93\% & 58.06\% & \cellcolor{gray!10}\textbf{88.31\%} {\tiny \textcolor{blue}{+20.89\%}} \\

& & Efficiency Score & 0.442 & \textbf{0.448} & 0.316 & 0.340 & \cellcolor{gray!10}0.405 {\tiny \textcolor{blue}{-9.60\%}} \\

\midrule

\multirow{8}{*}{\shortstack{Llama-3.1-\\8B-Instruct}} & \multirow{4}{*}{\shortstack{Paladin \\ Eval}} & Recovery Rate & 21.83\% & 63.44\% & 56.32\% & 49.2\% & \cellcolor{gray!10}\textbf{79.77\%} {\tiny \textcolor{blue}{+16.33\%}} \\

& & Task Success Rate & 17.46\% & 68.92\% & 53.74\% & 47.26\% & \cellcolor{gray!10}\textbf{78.72\%} {\tiny \textcolor{blue}{+9.80\%}} \\

& & Catastrophic Success Rate & 17.58\% & 71.84\% & 67.88\% & 65.47\% & \cellcolor{gray!10}\textbf{80.73\%} {\tiny \textcolor{blue}{+8.89\%}} \\

& & Efficiency Score &  \textbf{0.427} & 0.254 & 0.287 & 0.209 & \cellcolor{gray!10}0.323 {\tiny \textcolor{blue}{-24.36\%}} \\

\cmidrule(lr){2-8}

& \multirow{4}{*}{\shortstack{ToolReflect\\Eval }} & Recovery Rate & 18.32\% & 58.55\% & 49.32\% & 42.23\% & \cellcolor{gray!10}\textbf{73.34\%} {\tiny \textcolor{blue}{+14.79\%}} \\

& & Task Success Rate & 13.45\% & 59.47\% & 46.34\% & 41.22\% & \cellcolor{gray!10}\textbf{71.27\%} {\tiny \textcolor{blue}{+11.80\%}} \\

& & Catastrophic Success Rate & 15.40\% & 63.81\% & 58.09\% & 53.68\% & \cellcolor{gray!10}\textbf{71.77\%} {\tiny \textcolor{blue}{+7.96\%}} \\

& & Efficiency Score & \textbf{0.568} & 0.360 & 0.508 & 0.412 & \cellcolor{gray!10}0.385 {\tiny \textcolor{blue}{-32.31\%}} \\

\bottomrule

\end{tabular}
% \end{adjustbox}
% \end{sc}
\end{scriptsize}
\end{center}
\vskip -0.20in
\end{table*}
\subsection{Overall Performance}

Table~\ref{tab:results} summarizes results across 8 model–dataset pairs. PALADIN delivers consistent improvements on Recovery Rate (RR), Task Success Rate (TSR), and Catastrophic Success Rate (CSR), while incurring modest efficiency tradeoffs. Specifically, PALADIN achieves the highest RR in 7/8 settings (avg. +13.6\%, 95\% CI: [11.9, 15.2]), top TSR in 6/8 (avg. +10.2\%, CI: [8.4, 11.7]), and best CSR in 6/8 (avg. +9.2\%, CI: [7.5, 10.6]). All gains are statistically significant ($p<0.01$, bootstrap $N=1000$). 

Efficiency drops are measurable but bounded: on LLaMA-3.1-8B, Efficiency decreases by $0.183$, while on Qwen-2.5-14B it improves slightly (+0.012). These results situate PALADIN close to the Pareto frontier: no baseline achieves simultaneously higher safety and higher efficiency.

Correlation analyses reinforce these findings. Recovery and TSR are strongly correlated ($r=0.91$, $p<0.001$), confirming recovery as the direct driver of task completion. CSR and Efficiency exhibit a significant negative correlation ($r=-0.72$, CI: [$-0.79$, $-0.63$]), highlighting that retry-driven safety comes at modest compute cost. TSR shows negligible correlation with Efficiency ($r=0.12$), underscoring that efficiency-centric metrics alone are insufficient proxies for reliability.

\begin{figure}[!htbp]
    \centering
    \begin{minipage}{0.48\linewidth}
        \centering
        \includegraphics[width=\linewidth]{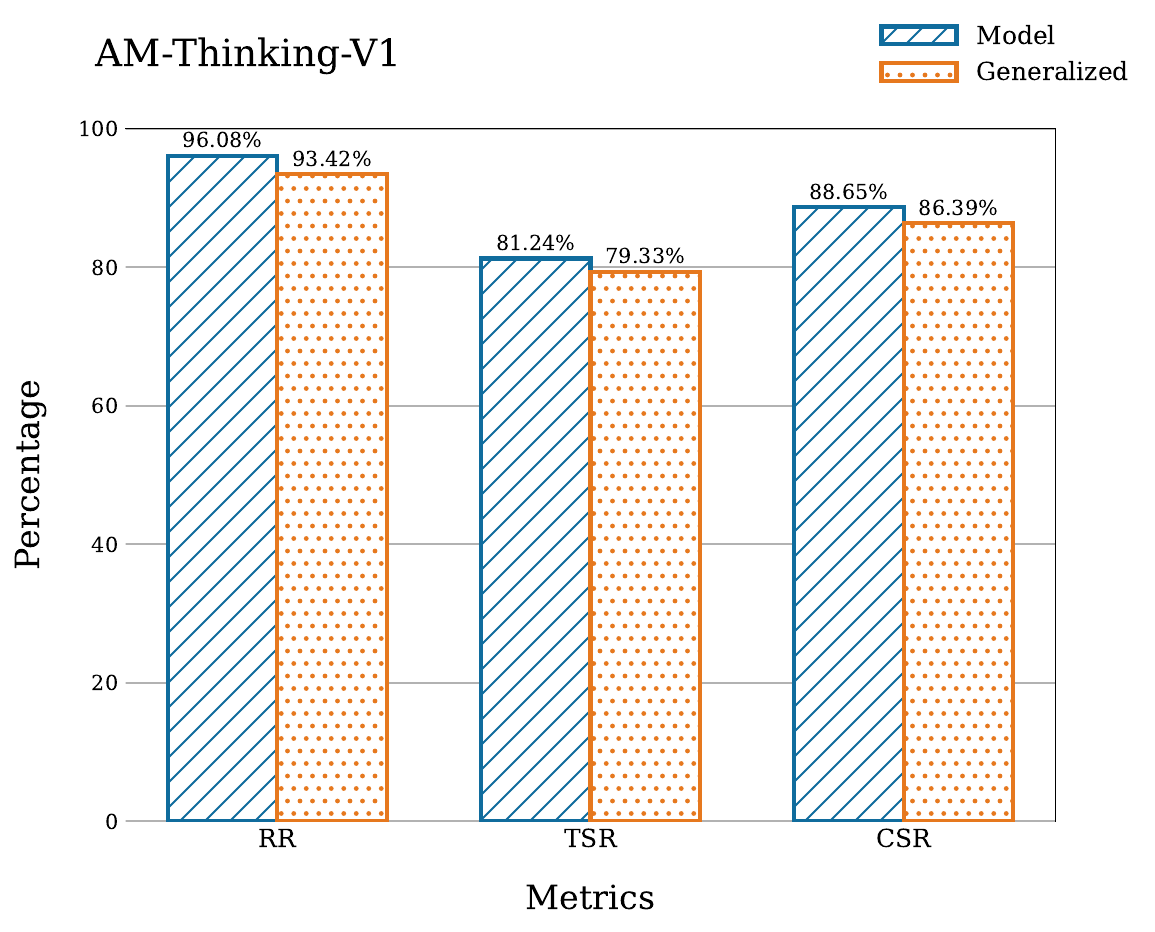}
        \label{fig:generatisation-amthinking}
    \end{minipage}\hfill
    \begin{minipage}{0.48\linewidth}
        \centering
        \includegraphics[width=\linewidth]{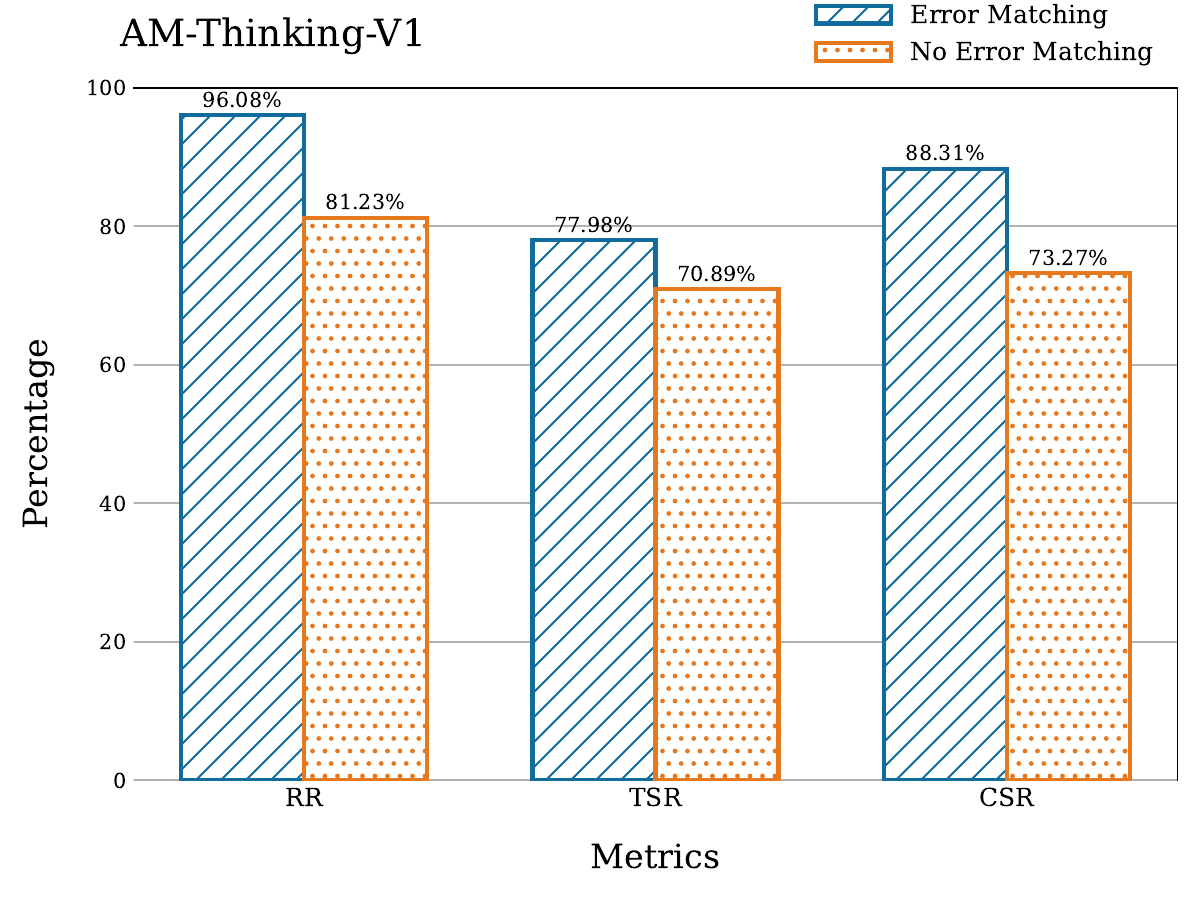}
        \label{fig:Abalation-amthinking}
    \end{minipage}
    \caption{PALADIN’s performance without inference time error matching compared to baseline across Gemma, Qwen, LLaMA, and AM-Thinking backbones. Refer Figure~\ref{fig:Abalation-amthinking} for full ablation graphs and Figure~\ref{fig:generalization} for full generalization graphs.}
    \label{fig:bargraphs}
\end{figure}

\subsection{Generalization}

Figures~\ref{fig:bargraphs} (a–d) emphasizes PALADIN’s generalizability across errors that it has never seen before. Despite differences in architecture and blind conditions, PALADIN consistently secures $>79\%$ RR, $>78\%$ TSR, and $>80\%$ CSR.

LLaMA-3.1-8B shows substantial improvement, with recovery rate (RR) rising from 21.8\% to 79.8\%, task success rate (TSR) increasing from 17.5\% to 78.7\%, and critical success rate (CSR) improving by 63.2 percentage points (from 17.6\% to 80.7\%). Qwen-2.5-14B achieves the highest CSR at 94.6\%, with RR at 94.7\% and TSR at 79.5\%, reducing silent failure cases nearly threefold compared to ToolReflect. Gemma-3-12B demonstrates balanced gains, with RR of 89.7\%, TSR of 87.4\%, and CSR of 82.6\%. AM-Thinking V1 records the highest RR overall at 96.1\%, alongside TSR of 81.2\% and CSR of 88.7\%.

\subsection{Ablations}
We ablate PALADIN’s inference-time recovery retrieval (Figures~\ref{fig:bargraphs}). Removing exemplar matching sharply reduces robustness.
Gemma-12B experiences a drop in performance, with recovery rate (RR) declining from 89.7\% to 61.4\%, task success rate (TSR) from 87.4\% to 57.3\%, and critical success rate (CSR) from 82.6\% to 65.1\%. Qwen-14B shows similar degradation, with RR decreasing from 94.7\% to 73.3\%, TSR from 79.5\% to 66.3\%, and CSR from 94.6\% to 68.9\%. LLaMA-8B sees RR fall from 79.8\% to 48.6\%, TSR from 78.7\% to 42.7\%, and CSR from 80.7\% to 57.4\%. AM-Thinking V1 maintains relatively higher robustness, with RR reducing from 96.1\% to 81.2\%, TSR from 81.2\% to 70.9\%, and CSR from 88.7\% to 73.3\%.
This confirms PALADIN’s robustness arises from both training-time exposure to failures and inference-time retrieval, with retrieval contributing up to +20–30 points in robustness metrics.

\subsection{Key Takeaways}
PALADIN achieves frontier-level robustness across backbones: large gains in recovery and safety, modest efficiency costs, and consistent generalization. Recovery supervision transforms brittle agents into reliable ones, proving that robustness is a learnable signal rather than an emergent artifact of scale. The ablation confirms retrieval is a critical ingredient, validating PALADIN’s design as both a training and inference-time intervention (see Appendix~\ref{app:metrics-ablations} for more details).

\section{Related Works}
\subsection{Execution-Level Robustness}
Recent methods look to harden tool-augmented agents against execution-time failures, but they are different in how they represent and learn from failure states. 
ToolReflect~\citep{polyakov-etal-2025-toolreflection} uses SFT on paired "bad vs. good" tool calls to teach models how to improve weak calls; while this yields a better call quality, it mainly shows the difference between single-call corrections instead of modeling a multi-step recovery path from cascading failures across a full trajectory. Recent work on LLM runtime error handling~\citep{sun2024llmruntimeerrorhandler} and structured exception handling~\citep{zhou2025shieldastructuredhandlingexceptions} has explored systematic approaches to agent resilience.
Critic-style approaches (Appendix~\ref{app:critic}) add in an evaluator that critiques and revises an agent’s actions, which improves reliability through iterative self-correction; but, these feedback loops happen reactively and are sometimes of poor quality, especially without diverse tool-error taxonomies. 
In contrast, PALADIN targets the robustness of execution by training directly on failure-rich trajectories that include diagnosis, replanning, and multi-turn recovery over multiple tool calls, rather than isolated “bad vs. good” pairs. While both Tool-Reflect and critic-style approaches demonstrate some gain in robustness, their emphasis is on post-hoc correction or call-level contrast rather than learned, trajectory-level recovery across varied execution failures.

\subsection{Diagnostics and Benchmarks}
Recent efforts in error diagnostics, such as ToolScan or BugGen, have similar motivations but are different in both scope and implementation. 
ToolScan~\citep{kokane2025toolscanbenchmarkcharacterizingerrors} provides a taxonomy of tool-use failures observed in LLMs, identifying seven recurring error patterns: Tool Hallucination, Argument Hallucination, Invalid Tool Invocation, Partial Tool Execution, Tool Output Hallucination, Invalid Intermediate Reasoning, and Re-entrant Error Handling Failures (see Appendix~\ref{app:error_t}). While it is valuable for understanding the variety of failures an agent can face, ToolScan is a diagnostic; meaning it labels and categorizes error modes rather than curating executable recovery paths or training agents on full recovery trajectories. 
BugGen, by contrast, emphasizes that handcrafted or mutation bug-synthesis tends to lead to narrow, weak coverage of real world failures, and they argue that realistic failure-rich datasets better reflect faults in the wild. Looked at together, ToolScan grounds the taxonomy of observed errors, whereas BugGen motivates realistic data generation and evaluation that capture real operational failures rather than synthetic failures.

\subsection{Prior Tool Systems \& Interface Reliability}
Foundational works like Toolformer~\citep{schick2023toolformer}, Gorilla~\citep{patil2023gorilla}, ToolLLM~\citep{qin2023toolllm}, ToolBench~\citep{guo2024stabletoolbench}, and RoTBench advanced agent capabilities across selection, multi-step planning, and formatting, showcasing strong performances under clean, benchmark-controlled environments and establishing multiple core principles for agentic tool usage. However, a common limitation found in most of these models is their dependence on ``happy-path'' trajectories – tool calls that assume ideal conditions in a model’s execution. 

Even when errors are present in the data, prior systems often lack explicit recovery strategies, leading to brittle behavior during real-world deployment. ToolBench provides extensive tool access data but has no recovery protocols in the face of an error, while RoTBench advances robustness evaluation by introducing noisy documentation and ambiguous tool specifications. Complementary tools such as ToolFuzz~\citep{milev2025toolfuzzautomatedagent} (see Appendix~\ref{app:toolf}) address pre-execution reliability by detecting documentation–tool mismatches via fuzz testing; while valuable for interface quality, this is not needed for runtime resilience, which requires detecting, diagnosing, and recovering from failures that arise during actual tool interaction.

\section{Discussion}

Our results yield four central insights.  
First, recovery is \emph{learnable}: PALADIN consistently converts failures into successes, raising TSR by more than fourfold on mid-scale models (17.5\% to 78.7\% on LLaMA-8B).  
Second, robustness is \emph{transferable}: across Gemma, Qwen, LLaMA, and AM-Thinking, PALADIN secures $>79\%$ RR, showing recovery supervision is a universal training signal rather than an emergent feature of scale.  
Third, robustness entails a quantifiable \emph{tradeoff}: retry intensity lowers catastrophic failures (82.4\% $\to$ 19.3\%) while modestly reducing efficiency, situating PALADIN near the Pareto frontier where no baseline dominates on both safety and cost.  
Fourth, recovery is \emph{retrieval-dependent}: our ablation shows inference-time exemplar matching contributes up to +30 points in robustness.
These insights shift the evaluation paradigm. Efficiency alone is a dangerous proxy: our correlation analysis ($r=-0.72$ between CSR and Efficiency) shows that models optimized purely for speed are systematically brittle. Robustness must instead be treated as a first-class objective—on par with accuracy and efficiency—if agents are to be deployed in failure-prone settings.  

PALADIN’s improvements stem not from scale or external reflection, but from explicit exposure to failures and proactive error handling. The targeted nature of its retries (fewer than one additional step on average) shows resilience does not require brute-force exploration, but data-driven learning of minimal corrective actions.  

\section{Conclusion}

PALADIN demonstrates that execution-level robustness is not emergent but a \emph{teachable, scalable capability}. By unifying recovery-rich fine-tuning with inference-time exemplar retrieval, it transforms brittle, “happy-path” agents into resilient problem-solvers. Across backbones, PALADIN achieves $>79\%$ RR, $>78\%$ TSR, and $>80\%$ CSR, while keeping efficiency penalties within one step.  
These contributions extend beyond benchmarks. PALADIN shows recovery behaviors generalize to unseen failures, proving robustness is not tied to specific error types but can be abstracted into transferable policies.  

The implications are twofold. For research, robustness supervision provides a new axis for evaluation and learning, complementary to scaling and alignment. For practice, PALADIN enables safer deployment of LLM agents in high-stakes domains where silent failures are unacceptable. 
Future work should explore adaptive controllers that modulate retry intensity based on task difficulty or model confidence, and integration with error logs from production environments. 

In short, PALADIN sets a new benchmark for execution-level resilience: robustness can be systematically taught, generalized across models, and achieved without prohibitive cost—laying the groundwork for safe, failure-aware AI systems.

\clearpage
\nocite{*}
\bibliography{main}

\begin{thebibliography}{70}
\providecommand{\natexlab}[1]{#1}
\providecommand{\url}[1]{\texttt{#1}}
\expandafter\ifx\csname urlstyle\endcsname\relax
  \providecommand{\doi}[1]{doi: #1}\else
  \providecommand{\doi}{doi: \begingroup \urlstyle{rm}\Url}\fi

\bibitem[Armbrust et~al.(2010)Armbrust, Fox, Griffith, Joseph, Katz, and et~al.]{armbrust2010view}
Michael Armbrust, Armando Fox, Rean Griffith, Anthony~D. Joseph, Randy Katz, and Andy~Konwinski et~al.
\newblock A view of cloud computing, 2010.

\bibitem[Avizienis et~al.(2004)Avizienis, Laprie, Randell, and Landwehr]{avizienis2004basic}
Algirdas Avizienis, Jean-Claude Laprie, Brian Randell, and Carl Landwehr.
\newblock Basic concepts and taxonomy of dependable and secure computing, 2004.

\bibitem[Biderman et~al.(2024)Biderman, Portes, Ortiz, Paul, Greengard, Jennings, King, et~al.]{biderman2024lora}
Dan Biderman, Jacob Portes, Jose Javier~Gonzalez Ortiz, Mansheej Paul, Philip Greengard, Connor Jennings, Daniel King, et~al.
\newblock Lora learns less and forgets less, 2024.
\newblock URL \url{https://arxiv.org/abs/2405.09673}.

\bibitem[Bray(2017)]{bray2017javascript}
Tim Bray.
\newblock The javascript object notation (json) data interchange format, 2017.
\newblock URL \url{https://tools.ietf.org/html/rfc8259}.

\bibitem[Chen et~al.(2025)Chen, Wang, Wu, Chen, Xu, Luo, et~al.]{chen2025advancing}
Sijia Chen, Yibo Wang, Yi-Feng Wu, Qing-Guo Chen, Zhao Xu, Weihua Luo, et~al.
\newblock Advancing tool-augmented large language models: Integrating insights from errors in inference trees, 2025.
\newblock URL \url{https://arxiv.org/abs/2406.07115}.

\bibitem[Christiano et~al.(2017)Christiano, Leike, Brown, Martic, Legg, and Amodei]{christiano2017deep}
Paul~F. Christiano, Jan Leike, Tom Brown, Miljan Martic, Shane Legg, and Dario Amodei.
\newblock Deep reinforcement learning from human preferences, 2017.
\newblock URL \url{https://arxiv.org/abs/1706.03741}.

\bibitem[Cormen et~al.(2009)Cormen, Leiserson, Rivest, and Stein]{cormen2009introduction}
Thomas~H. Cormen, Charles~E. Leiserson, Ronald~L. Rivest, and Clifford Stein.
\newblock Introduction to algorithms, 2009.

\bibitem[Coulouris et~al.(2011)Coulouris, Dollimore, Kindberg, and Blair]{coulouris2011distributed}
George Coulouris, Jean Dollimore, Tim Kindberg, and Gordon Blair.
\newblock Distributed systems: Concepts and design, 2011.

\bibitem[Dettmers et~al.(2023)Dettmers, Pagnoni, Holtzman, and Zettlemoyer]{dettmers2023qlora}
Tim Dettmers, Artidoro Pagnoni, Ari Holtzman, and Luke Zettlemoyer.
\newblock Qlora: Efficient finetuning of quantized llms, 2023.
\newblock URL \url{https://arxiv.org/abs/2305.14314}.

\bibitem[Driess et~al.(2023)Driess, Xia, Sajjadi, Lynch, Chowdhery, Ichter, et~al.]{driess2023palm}
Danny Driess, Fei Xia, Mehdi S.~M. Sajjadi, Corey Lynch, Aakanksha Chowdhery, Brian Ichter, et~al.
\newblock Palm-e: An embodied multimodal language model, 2023.
\newblock URL \url{https://arxiv.org/abs/2303.03378}.

\bibitem[Feng et~al.(2024)Feng, He, Huang, Lin, Zhang, Zhang, and Li]{feng2024agile}
Peiyuan Feng, Yichen He, Guanhua Huang, Yuan Lin, Hanchong Zhang, Yuchen Zhang, and Hang Li.
\newblock Agile: A novel reinforcement learning framework of llm agents, 2024.
\newblock URL \url{https://arxiv.org/abs/2405.14751}.

\bibitem[Fielding \& Reschke(2014)Fielding and Reschke]{fielding2014http}
Roy Fielding and Julian Reschke.
\newblock Hypertext transfer protocol (http/1.1): Semantics and content, 2014.
\newblock URL \url{https://tools.ietf.org/html/rfc7231}.

\bibitem[Grattafiori et~al.(2024)Grattafiori, Dubey, Jauhri, Pandey, Kadian, Al-Dahle, et~al.]{dubey2024llama3}
Aaron Grattafiori, Abhimanyu Dubey, Abhinav Jauhri, Abhinav Pandey, Abhishek Kadian, Ahmad Al-Dahle, et~al.
\newblock The llama 3 herd of models, 2024.
\newblock URL \url{https://arxiv.org/abs/2407.21783}.

\bibitem[Gray(1985)]{gray1985measurement}
Jim Gray.
\newblock Measurement of availability and recovery in computer systems, 1985.

\bibitem[Guo et~al.(2024)Guo, Cheng, Wang, Liang, Qin, Li, et~al.]{guo2024stabletoolbench}
Zhicheng Guo, Sijie Cheng, Hao Wang, Shihao Liang, Yujia Qin, Peng Li, et~al.
\newblock Stabletoolbench: Towards stable large-scale benchmarking on tool learning of large language models, 2024.
\newblock URL \url{https://arxiv.org/abs/2403.07714}.

\bibitem[Hammond et~al.(2025)Hammond, Chan, Clifton, Hoelscher-Obermaier, Khan, Khan, et~al.]{hammond2025multiagent}
Lewis Hammond, Alan Chan, Jesse Clifton, Jason Hoelscher-Obermaier, Akbir Khan, Akbir Khan, et~al.
\newblock Multi-agent risks from advanced ai, 2025.
\newblock URL \url{https://arxiv.org/abs/2502.14143}.

\bibitem[Hendrycks et~al.(2020)Hendrycks, Burns, Basart, Zou, Mazeika, Song, and Steinhardt]{hendrycks2020measuring}
Dan Hendrycks, Collin Burns, Steven Basart, Andy Zou, Mantas Mazeika, Dawn Song, and Jacob Steinhardt.
\newblock Measuring massive multitask language understanding, 2020.
\newblock URL \url{https://arxiv.org/abs/2009.03300}.

\bibitem[Hu et~al.(2021)Hu, Shen, Wallis, Allen-Zhu, Li, Wang, et~al.]{hu2021lora}
Edward~J. Hu, Yelong Shen, Phillip Wallis, Zeyuan Allen-Zhu, Yuanzhi Li, Shean Wang, et~al.
\newblock Lora: Low-rank adaptation of large language models, 2021.
\newblock URL \url{https://arxiv.org/abs/2106.09685}.

\bibitem[Jacob et~al.(2017)Jacob, Kligys, Chen, Zhu, Tang, Howard, Adam, and Kalenichenko]{jacob2017quantization}
Benoit Jacob, Skirmantas Kligys, Bo~Chen, Menglong Zhu, Matthew Tang, Andrew Howard, Hartwig Adam, and Dmitry Kalenichenko.
\newblock Quantization and training of neural networks for efficient integer-arithmetic-only inference, 2017.
\newblock URL \url{https://arxiv.org/abs/1712.05877}.

\bibitem[Jasper et~al.(2025)Jasper, Luu, Pan, Tyagi, Quinn, Hu, and Houngninou]{jasper2025buggen}
Surya Jasper, Minh Luu, Evan Pan, Aakash Tyagi, Michael Quinn, Jiang Hu, and David~Kebo Houngninou.
\newblock Buggen: A self-correcting multi-agent llm pipeline for realistic rtl bug synthesis, 2025.
\newblock URL \url{https://arxiv.org/abs/2506.10501}.

\bibitem[Kokane et~al.(2025)Kokane, Zhu, Awalgaonkar, Zhang, Hoang, Prabhakar, Liu, Lan, Yang, Tan, Murthy, Yao, Liu, Niebles, Wang, Heinecke, Xiong, and Savarese]{kokane2025toolscanbenchmarkcharacterizingerrors}
Shirley Kokane, Ming Zhu, Tulika Awalgaonkar, Jianguo Zhang, Thai Hoang, Akshara Prabhakar, Zuxin Liu, Tian Lan, Liangwei Yang, Juntao Tan, Rithesh Murthy, Weiran Yao, Zhiwei Liu, Juan~Carlos Niebles, Huan Wang, Shelby Heinecke, Caiming Xiong, and Silivo Savarese.
\newblock Toolscan: A benchmark for characterizing errors in tool-use llms, 2025.
\newblock URL \url{https://arxiv.org/abs/2411.13547}.

\bibitem[Korf(1985)]{korf1985depth}
Richard~E. Korf.
\newblock Depth-first iterative-deepening: An optimal admissible tree search, 1985.

\bibitem[Lambert(2025)]{lambert2024reinforcement}
Nathan Lambert.
\newblock Reinforcement learning from human feedback, 2025.
\newblock URL \url{https://arxiv.org/abs/2504.12501}.

\bibitem[Laprie(1992)]{laprie1992dependability}
Jean-Claude Laprie.
\newblock Dependability: Basic concepts and terminology, 1992.

\bibitem[Lin(2004)]{lin2004rouge}
Chin-Yew Lin.
\newblock Rouge: A package for automatic evaluation of summaries, 2004.

\bibitem[Liu et~al.(2023)Liu, Yu, Zhang, Yang, Lei, Lai, et~al.]{liu2023agentbench}
Xiao Liu, Hao Yu, Hanchen Zhang, Yifan Yang, Xuanyu Lei, Hanyu Lai, et~al.
\newblock Agentbench: Evaluating llms as agents, 2023.
\newblock URL \url{https://arxiv.org/abs/2308.03688}.

\bibitem[Liu et~al.(2024)Liu, Lyn, Zhu, Tian, and Graham]{zhu2024alora}
Zequan Liu, Jiawen Lyn, Wei Zhu, Xing Tian, and Yvette Graham.
\newblock Alora: Allocating low-rank adaptation for fine-tuning large language models, 2024.
\newblock URL \url{https://arxiv.org/abs/2403.16187}.

\bibitem[Ma et~al.(2024)Ma, Huang, , Zhang, Gypta, and Krishna]{ma2024mm}
Zixian Ma, Weikai Huang, , Jieyu Zhang, Tanmay Gypta, and Ranjay Krishna.
\newblock m\&m's: A benchmark to evaluate tool-use for multi-step multi-modal tasks, 2024.
\newblock URL \url{https://arxiv.org/abs/2403.11085}.

\bibitem[Madaan et~al.(2023)Madaan, Tandon, Gupta, Hallinan, Gao, Wiegreffe, et~al.]{madaan2023selfrefine}
Aman Madaan, Niket Tandon, Prakhar Gupta, Skyler Hallinan, Luyu Gao, Sarah Wiegreffe, et~al.
\newblock Self-refine: Iterative refinement with self-feedback, 2023.
\newblock URL \url{https://arxiv.org/abs/2303.17651}.

\bibitem[Milev et~al.(2025)Milev, Balunović, Baader, and Vechev]{milev2025toolfuzzautomatedagent}
Ivan Milev, Mislav Balunović, Maximilian Baader, and Martin Vechev.
\newblock Toolfuzz -- automated agent tool testing, 2025.
\newblock URL \url{https://arxiv.org/abs/2503.04479}.

\bibitem[Mohammadi et~al.(2025)Mohammadi, Li, Lo, and Yip]{mohammadi2025evaluation}
Mahmoud Mohammadi, Yipeng Li, Jane Lo, and Wendy Yip.
\newblock Evaluation and benchmarking of llm agents: A survey, 2025.
\newblock URL \url{https://arxiv.org/abs/2507.21504}.

\bibitem[Nagel et~al.(2021)Nagel, Fournarakis, Amjad, Bondarenko, van Baalen, and Blankevoort]{nagel2021white}
Markus Nagel, Marios Fournarakis, Rana~Ali Amjad, Yelysei Bondarenko, Mart van Baalen, and Tijmen Blankevoort.
\newblock A white paper on neural network quantization, 2021.
\newblock URL \url{https://arxiv.org/abs/2106.08295}.

\bibitem[Network(2024)]{mozilla2024http}
Mozilla~Developer Network.
\newblock Http response status codes, 2024.
\newblock URL \url{https://developer.mozilla.org/en-US/docs/Web/HTTP/Status}.

\bibitem[Nottingham et~al.(2023)Nottingham, Wilde, and Dalal]{nottingham2023problem}
Mark Nottingham, Erik Wilde, and Sanjay Dalal.
\newblock Problem details for http apis, 2023.
\newblock URL \url{https://tools.ietf.org/html/rfc9457}.

\bibitem[Ouyang et~al.(2022)Ouyang, Wu, Jiang, Almeida, Wainwright, Mishkin, et~al.]{ouyang2022training}
Long Ouyang, Jeffrey Wu, Xu~Jiang, Diogo Almeida, Carroll Wainwright, Pamela Mishkin, et~al.
\newblock Training language models to follow instructions with human feedback, 2022.
\newblock URL \url{https://arxiv.org/abs/2203.02155}.

\bibitem[Papineni et~al.(2002)Papineni, Roukos, Ward, and Zhu]{papineni2002bleu}
Kishore Papineni, Salim Roukos, Todd Ward, and Wei-Jing Zhu.
\newblock Bleu: A method for automatic evaluation of machine translation, 2002.

\bibitem[Patil et~al.(2023)Patil, Zhang, Wang, and Gonzalez]{patil2023gorilla}
Shishir~G. Patil, Tianjun Zhang, Xin Wang, and Joseph~E. Gonzalez.
\newblock Gorilla: Large language model connected with massive apis, 2023.
\newblock URL \url{https://arxiv.org/abs/2305.15334}.

\bibitem[Pezoa et~al.(2016)Pezoa, Reutter, Suarez, Ugarte, and Vrgoč]{pezoa2016foundations}
Felipe Pezoa, Juan~L. Reutter, Fernando Suarez, Martín Ugarte, and Domagoj Vrgoč.
\newblock Foundations of json schema, 2016.

\bibitem[Polyakov et~al.(2025)Polyakov, Alimova, Abulkhanov, Sedykh, Bout, Nikolenko, and Piontkovskaya]{polyakov-etal-2025-toolreflection}
Gregory Polyakov, Ilseyar Alimova, Dmitry Abulkhanov, Ivan Sedykh, Andrey Bout, Sergey Nikolenko, and Irina Piontkovskaya.
\newblock {T}ool{R}eflection: Improving large language models for real-world {API} calls with self-generated data.
\newblock In Ehsan Kamalloo, Nicolas Gontier, Xing~Han Lu, Nouha Dziri, Shikhar Murty, and Alexandre Lacoste (eds.), \emph{Proceedings of the 1st Workshop for Research on Agent Language Models (REALM 2025)}, pp.\  184--199, Vienna, Austria, July 2025. Association for Computational Linguistics.
\newblock ISBN 979-8-89176-264-0.
\newblock \doi{10.18653/v1/2025.realm-1.14}.
\newblock URL \url{https://aclanthology.org/2025.realm-1.14/}.

\bibitem[Qin et~al.(2023)Qin, Liang, Ye, Zhu, Yan, Lu, et~al.]{qin2023toolllm}
Yujia Qin, Shihao Liang, Yining Ye, Kunlun Zhu, Lan Yan, Yaxi Lu, et~al.
\newblock Toolllm: Facilitating large language models to master 16000+ real-world apis, 2023.
\newblock URL \url{https://arxiv.org/abs/2307.16789}.

\bibitem[Qu et~al.(2024)Qu, Dai, Wei, Cai, Wang, Yin, et~al.]{qu2024toollearning}
Changle Qu, Sunhao Dai, Xiaochi Wei, Hengyi Cai, Shuaiqiang Wang, Dawei Yin, et~al.
\newblock Tool learning with large language models: A survey, 2024.
\newblock URL \url{https://arxiv.org/abs/2405.17935}.

\bibitem[Rafailov et~al.(2023)Rafailov, Sharma, Mitchell, Ermon, Manning, and Finn]{rafailov2023direct}
Rafael Rafailov, Archit Sharma, Eric Mitchell, Stefano Ermon, Christopher~D. Manning, and Chelsea Finn.
\newblock Direct preference optimization: Your language model is secretly a reward model, 2023.
\newblock URL \url{https://arxiv.org/abs/2305.18290}.

\bibitem[Russell \& Norvig(2020)Russell and Norvig]{russell2020artificial}
Stuart~J. Russell and Peter Norvig.
\newblock Artificial intelligence: A modern approach, 2020.

\bibitem[Sane(2025)]{sane2025amppoppo}
Soham Sane.
\newblock Am-ppo: (advantage) alpha-modulation with proximal policy optimization, 2025.
\newblock URL \url{https://arxiv.org/abs/2505.15514}.

\bibitem[Schick et~al.(2023)Schick, Dwivedi-Yu, Dessì, Raileanu, Lomeli, Zettlemoyer, Cancedda, and Scialom]{schick2023toolformer}
Timo Schick, Jane Dwivedi-Yu, Roberto Dessì, Roberta Raileanu, Maria Lomeli, Luke Zettlemoyer, Nicola Cancedda, and Thomas Scialom.
\newblock Toolformer: Language models can teach themselves to use tools, 2023.
\newblock URL \url{https://arxiv.org/abs/2302.04761}.

\bibitem[Schulman et~al.(2017)Schulman, Wolski, Dhariwal, Radford, and Klimov]{schulman2017proximalpolicyoptimization}
John Schulman, Filip Wolski, Prafulla Dhariwal, Alec Radford, and Oleg Klimov.
\newblock Proximal policy optimization algorithms, 2017.
\newblock URL \url{https://arxiv.org/abs/1707.06347}.

\bibitem[Services(2024{\natexlab{a}})]{aws2024retry}
Amazon~Web Services.
\newblock Retry with backoff pattern - aws prescriptive guidance, 2024{\natexlab{a}}.
\newblock URL \url{https://docs.aws.amazon.com/prescriptive-guidance/latest/cloud-design-patterns/retry-backoff.html}.

\bibitem[Services(2024{\natexlab{b}})]{aws2024timeouts}
Amazon~Web Services.
\newblock Timeouts, retries and backoff with jitter - the amazon builders' library, 2024{\natexlab{b}}.
\newblock URL \url{https://aws.amazon.com/builders-library/timeouts-retries-and-backoff-with-jitter/}.

\bibitem[Sha et~al.(2025)Sha, Tian, Xu, Cui, Meng, and Wang]{sha2025agentsafety}
Zeyang Sha, Hanling Tian, Zhuoer Xu, Shiwen Cui, Changhua Meng, and Weiqiang Wang.
\newblock Agent safety alignment via reinforcement learning, 2025.
\newblock URL \url{https://arxiv.org/abs/2507.08270}.

\bibitem[Shen et~al.(2023)Shen, Song, Tan, Li, Lu, and Zhuang]{shen2023hugginggpt}
Yongliang Shen, Kaitao Song, Xu~Tan, Dongsheng Li, Weiming Lu, and Yueting Zhuang.
\newblock Hugginggpt: Solving ai tasks with chatgpt and its friends in hugging face, 2023.
\newblock URL \url{https://arxiv.org/abs/2303.17580}.

\bibitem[Singh et~al.(2025)Singh, Magazine, Pandya, and Nambi]{singh2025artist}
Joykirat Singh, Raghav Magazine, Yash Pandya, and Akshay Nambi.
\newblock Agentic reasoning and tool integration for llms via reinforcement learning, 2025.
\newblock URL \url{https://arxiv.org/abs/2505.01441}.

\bibitem[Stevens et~al.(2013)Stevens, Fenner, and Rudoff]{stevens2013unix}
W.~Richard Stevens, Bill Fenner, and Andrew~M. Rudoff.
\newblock Unix network programming, volume 1: The sockets networking api, 2013.

\bibitem[Sun et~al.(2025)Sun, Yang, Duan, Shi, Lyu, Cheng, et~al.]{sun2025multiagent}
Lijun Sun, Yijun Yang, Qiqi Duan, Yuhui Shi, Chao Lyu, Yu-Cheng Cheng, et~al.
\newblock Multi-agent coordination across diverse applications: A survey, 2025.
\newblock URL \url{https://arxiv.org/abs/2502.14743}.

\bibitem[Sun et~al.(2024)Sun, Zhu, Xu, Du, Li, and Lo]{sun2024llmruntimeerrorhandler}
Zhensu Sun, Haotian Zhu, Bowen Xu, Xiaoning Du, Li~Li, and David Lo.
\newblock Llm as runtime error handler: A promising pathway to adaptive self-healing of software systems, 2024.
\newblock URL \url{https://arxiv.org/abs/2408.01055}.

\bibitem[Tanenbaum \& Wetherall(2016)Tanenbaum and Wetherall]{tanenbaum2016computer}
Andrew~S. Tanenbaum and David~J. Wetherall.
\newblock Computer networks, 2016.

\bibitem[Touvron et~al.(2023{\natexlab{a}})Touvron, Lavril, Izacard, Martinet, Lachaux, Lacroix, et~al.]{touvron2023llama}
Hugo Touvron, Thibaut Lavril, Gautier Izacard, Xavier Martinet, Marie-Anne Lachaux, Timothée Lacroix, et~al.
\newblock Llama: Open and efficient foundation language models, 2023{\natexlab{a}}.
\newblock URL \url{https://arxiv.org/abs/2302.13971}.

\bibitem[Touvron et~al.(2023{\natexlab{b}})Touvron, Martin, Stone, Albert, Almahairi, Babaei, et~al.]{touvron2023llama2}
Hugo Touvron, Louis Martin, Kevin Stone, Peter Albert, Amjad Almahairi, Yasmine Babaei, et~al.
\newblock Llama 2: Open foundation and fine-tuned chat models, 2023{\natexlab{b}}.
\newblock URL \url{https://arxiv.org/abs/2307.09288}.

\bibitem[Wang et~al.(2024)Wang, Gao, Meng, Peng, Hu, Lin, and Gao]{wang2024aegis}
Xinchen Wang, Pengfei Gao, Xiangxin Meng, Chao Peng, Ruida Hu, Yun Lin, and Cuiyun Gao.
\newblock Aegis: An agent-based framework for general bug reproduction from issue descriptions, 2024.
\newblock URL \url{https://arxiv.org/abs/2411.18015}.

\bibitem[Wikipedia(2024)]{exponentialbackoff2024}
Wikipedia.
\newblock Exponential backoff, 2024.
\newblock URL \url{https://en.wikipedia.org/wiki/Exponential_backoff}.

\bibitem[Yang et~al.(2025)Yang, Chai, Song, Qi, Wen, Li, and otherse]{yang2025survey}
Yingxuan Yang, Huacan Chai, Yuanyi Song, Siyuan Qi, Muning Wen, Ning Li, and otherse.
\newblock A survey of ai agent protocols, 2025.
\newblock URL \url{https://arxiv.org/abs/2504.16736}.

\bibitem[Yao et~al.(2022)Yao, Zhao, Yu, Du, Shafran, Narasimhan, and Cao]{yao2022react}
Shunyu Yao, Jeffrey Zhao, Dian Yu, Nan Du, Izhak Shafran, Karthik Narasimhan, and Yuan Cao.
\newblock React: Synergizing reasoning and acting in language models, 2022.
\newblock URL \url{https://arxiv.org/abs/2210.03629}.

\bibitem[Yao et~al.(2023)Yao, Yu, Zhao, Shafran, Griffiths, Cao, and Narasimhan]{yao2022treeofthoughts}
Shunyu Yao, Dian Yu, Jeffrey Zhao, Izhak Shafran, Thomas~L. Griffiths, Yuan Cao, and Karthik Narasimhan.
\newblock Tree of thoughts: Deliberate problem solving with large language models, 2023.
\newblock URL \url{https://arxiv.org/abs/2305.10601}.

\bibitem[Ye et~al.(2024)Ye, Xu, Li, and Allen-Zhu]{ye2024physicslanguagemodels22}
Tian Ye, Zicheng Xu, Yuanzhi Li, and Zeyuan Allen-Zhu.
\newblock Physics of language models: Part 2.2, how to learn from mistakes on grade-school math problems, 2024.
\newblock URL \url{https://arxiv.org/abs/2408.16293}.

\bibitem[Yu et~al.(2023)Yu, Lee, Raghavan, Samarasekara, Tokekar, and Hare]{yu2023enhancing}
Peihong Yu, Bhoram Lee, Aswin Raghavan, Supuri Samarasekara, Pratap Tokekar, and James~Zachary Hare.
\newblock Enhancing multi-agent coordination through common operating picture integration, 2023.
\newblock URL \url{https://arxiv.org/abs/2311.04740}.

\bibitem[Yu et~al.(2025)Yu, Yang, Li, Zhang, Wang, Feng, et~al.]{yu2025multi}
Peijie Yu, Yifan Yang, Jinjian Li, Zelong Zhang, Haorui Wang, Xiao Feng, et~al.
\newblock Multi-mission tool bench: Assessing the robustness of llm based agents through related and dynamic missions, 2025.
\newblock URL \url{https://arxiv.org/abs/2504.02623}.

\bibitem[Zeng et~al.(2025)Zeng, Ding, Wang, Liu, Ning, Hou, et~al.]{zeng2025boosting}
Yirong Zeng, Xiao Ding, Yuxian Wang, Weiwen Liu, Wu~Ning, Yutai Hou, et~al.
\newblock itool: Reinforced fine-tuning with dynamic deficiency calibration for advanced tool use, 2025.
\newblock URL \url{https://arxiv.org/abs/2501.09766}.

\bibitem[Zhang et~al.(2025)Zhang, Wu, Yu, Lv, and Zhang]{zhang2025robustness}
Kun Zhang, Le~Wu, Kui Yu, Guangyi Lv, and Dacao Zhang.
\newblock Evaluating and improving robustness in large language models: A survey and future directions, 2025.
\newblock URL \url{https://arxiv.org/abs/2506.11111}.

\bibitem[Zhao et~al.(2025)Zhao, Yuksekgonul, Wu, and Zou]{zhao2025sirius}
Wanjia Zhao, Mert Yuksekgonul, Shirley Wu, and James Zou.
\newblock Sirius: Self-improving multi-agent systems via bootstrapped reasoning, 2025.
\newblock URL \url{https://arxiv.org/abs/2502.04780}.

\bibitem[Zheng et~al.(2023)Zheng, Dou, Gao, Hua, Shen, Wang, et~al.]{gao2023secrets}
Rui Zheng, Shihan Dou, Songyang Gao, Yuan Hua, Wei Shen, Binghai Wang, et~al.
\newblock Secrets of rlhf in large language models part i: Ppo, 2023.
\newblock URL \url{https://arxiv.org/abs/2307.04964}.

\bibitem[Zhou et~al.(2025)Zhou, Chen, Lu, Zhao, and Zhu]{zhou2025shieldastructuredhandlingexceptions}
Jingwen Zhou, Jieshan Chen, Qinghua Lu, Dehai Zhao, and Liming Zhu.
\newblock Shielda: Structured handling of exceptions in llm-driven agentic workflows, 2025.
\newblock URL \url{https://arxiv.org/abs/2508.07935}.

\end{thebibliography}
\bibliographystyle{iclr2026_conference}

\appendix
\clearpage
\appendix{\large{\textbf{Appendix}}}
\section{Critic-Style Approach}
\label{app:critic}
We implemented a critic-style agent baseline with an oracle-assisted loop as a benchmark for comparison to PALADIN. The critic module (Gemma-12B) was invoked after each tool call to detect execution errors and propose recovery actions. Whenever an error was flagged, the executor was directed to an oracle dataset with pre-defined recovery actions for the detected error type. The most appropriate action among them was then selected to continue the trajectory. A maximum of 3 recovery attempts were permitted per error before execution continued with the latest attempt or failed gracefully.

Our implementation differs from prior CRITIC systems in that, rather than requiring the model to autonomously generate new candidate recovery actions or perform external reasoning (e.g., web search), we provide a structured set of recovery actions from a curated oracle dataset. The model’s role was to choose the most appropriate recovery, rather than generate one from scratch.
\subsection*{Evaluation Example: ServiceDependencyFailure}
\textbf{Chosen error:} ServiceDependencyFailure

\textbf{Justification:} The request involves two services: fetching hot products from AliExpress and validating email domains. A realistic failure is that one upstream service is temporarily unavailable. The model must recognize this is an external failure, not due to malformed input.

\textbf{Expected recovery actions:} Identify the failing service, wait until it is healthy, and retry the request.

\section{ToolFuzz}
\label{app:toolf}
While PALADIN focuses on runtime robustness during tool execution, complementary work such as ToolFuzz addresses pre-execution reliability by improving the alignment between tool documentation and model expectations.

ToolFuzz applies automated fuzz testing to API schemas in order to detect inconsistencies between declared documentation and actual behavior. It uncovered over 20x more specification-related failures than prompt engineering baselines across 32 LangChain and 35 custom tools—revealing widespread documentation underspecification as a root cause of tool-use errors.

Although PALADIN assumes that documentation is accurate at test time (as each tool is provided with correct instructions), ToolFuzz supports our broader vision: robust real-world tool use requires not only runtime adaptability, but also upstream validation of the tool interfaces themselves. We view ToolFuzz as complementary infrastructure—ensuring that PALADIN's recovery logic is exercised on meaningful failures, not avoidable documentation bugs.

\section{Runtime Errors}
\label{app:runtime_e}
\clearpage
\begin{table}[t]
\centering
\scriptsize
\renewcommand{\arraystretch}{1.25}
\setlength{\tabcolsep}{6pt}
\begin{tabular}{
    p{2.3cm} 
    >{\raggedright\arraybackslash}p{3cm} 
    >{\raggedright\arraybackslash}p{3.7cm} 
    >{\raggedright\arraybackslash}p{3.7cm}
}
\toprule
\textbf{Specific Failure} & 
\textbf{Example (Simulated Output)} & 
\textbf{Corrective Action Policy} & 
\textbf{Rationale and Citations} \\
\midrule
400 Bad Request & 
\ttfamily "error": "Malformed request syntax", "status": 400 & 
Re-examine tool documentation, check parameter formatting, and re-issue the call. & 
The error originates from the client’s request. The only path to recovery is for the client (the agent) to correct its own mistake. \\
\midrule
401 Unauthorized & 
\ttfamily "error": "Invalid API key provided", "status": 401 & 
Check for a valid API key. If missing or invalid, terminate the task and report the failure. & 
A non-recoverable authentication error; retrying with invalid credentials is futile. \\
\midrule
403 Forbidden & 
\ttfamily "error": "User does not have permission for this resource", "status": 403 & 
Terminate the task and report the lack of permissions. Do not retry. & 
This is an authorization failure. Retrying will not change permissions. \\
\midrule
404 Not Found & 
\ttfamily "error": "The requested resource does not exist", "status": 404 & 
Verify the request URL. If correct, assume the resource is unavailable and try an alternative tool. & 
A common error caused by typos or moved resources. Check self-error first. \\
\midrule
500 Internal Server Error & 
\ttfamily "error": "Unexpected server error", "status": 500 & 
Retry using exponential backoff. If failure persists after 3–4 attempts, terminate and report. & 
A catch-all for transient server-side issues. Retrying is industry standard. \\
\midrule
503 Service Unavailable & 
\ttfamily "error": "Service unavailable due to overload or maintenance", "status": 503 & 
Follow the Retry-After header if present; otherwise, use exponential backoff. & 
Retry is expected, as the issue is temporary. \\
\midrule
Request Timeout & 
\ttfamily \seqsplit{requests.exceptions.Timeout} & 
Retry with exponential backoff; distinguish connection vs. read timeouts. & 
Timeouts are usually transient and should be retried. \\
\midrule
DNS Resolution Error & 
\ttfamily \seqsplit{requests.exceptions.ConnectionError:getaddrinfo\_failed} & 
Check hostname for typos. If correct, wait and retry. & 
Could be a typo or a temporary DNS failure. \\
\midrule
429 Too Many Requests & 
\ttfamily "error": "Rate limit exceeded", "status": 429 & 
Respect the Retry-After header or apply exponential backoff. & 
Standard API rate enforcement. Ignoring leads to blocking. \\
\midrule
Malformed JSON & 
\ttfamily \seqsplit{SyntaxError:JSON. parse\_error} & 
Retry first. If repeated, use a lenient parser or fallback tool. & 
Often due to truncated or corrupted responses. Retry is the simplest fix. \\
\midrule
JSON Schema Violation & 
\ttfamily \seqsplit{ValidationError:'abc'is\_not\_of\_type'number'} & 
Report a data quality issue. Coerce if allowed, otherwise discard. & 
Data violates schema expectations. Must be handled gracefully. \\
\bottomrule
\end{tabular}
\caption{Catalog of common failures, simulated outputs, and recovery strategies.}
\label{tab:failures}
\end{table}
These are just some of the many common errors found in API faults, tool calls, or in daily life. For the rest of the errors used to train this model, check out our github: https://github.com/HexaA2/paladin/tree/main.

\clearpage
\section{Systematically Augmented Traces}
\label{app:traces}
Systematically augmented traces are clean ToolBench style agents–tool execution logs that are enriched with controlled, labeled failures and paired with different recovery trajectories, helping enable training that was reproducible and the evaluation of different recovery behaviors under realistic tool-level faults. This approach helped yield multiple deterministic variants of the same base rollout by injecting specific error types at different steps, with structured annotations for failure detection, diagnosis, and multi-turn correction.
Augmentations were generated by applying a failure-policy dictionary over baseline rollouts to inject errors such as timeouts, 5xx responses, rate limits, malformed tool outputs, and partial results, with configurable positions, frequencies, and cascades. Each injected failure was paired with an annotated recovery path (diagnose, replan, parameter fix, retry, or tool-switch), helping yield trajectory-level supervision for LoRA-based SFT and deterministic evaluation via metrics like TSR, RR, CSR, and ES. Because variants are deterministic, they support apples-to-apples ablations (e.g., single- vs multi-fault) and stable comparisons across model sizes and training regimes.
\begin{itemize}
    \item Base trace: plan → call Tool A with parameters p → receive valid response → continue.
    \item Augmented variant: identical plan → call Tool A with p → injected timeout at step t → agent detects timeout → backs off with jittered retry and parameter validation → successful response → proceed to completion. A second variant might inject a malformed JSON response instead of a timeout, prompting schema validation, selective re-query, or tool-switch. Each variant includes labels for failure type, step index, recovery rationale, and final outcome, enabling targeted training and evaluation.
\end{itemize}
\begin{lstlisting}[caption={Paladin trace with recovery and Finish}]
[{"role": "system","content": [{"type": "text","text": "You are **Paladin**, an error-resilient agent ..."}]},
{"role": "user","content": [{"type": "text","text": "I'm conducting a research project on social media influencers ..."}]},
{"role": "assistant","content": [{"type": "text","text": "Thought: I need to gather user information ... 
Action: userinfo_for_instagram_cheapest
Action Input: {
 \"username\": \"nike\"
}"}]},
{"role": "function","content": [{"type": "text","text": "{\"error\": \"\", \"response\": \"{'data': {'user': {'biography': 'Spotlighting athlete* ..."}]}]},
{"role": "assistant","content": [{"type": "text","text": "Recovery: Thought: The API call to retrieve user information ... 
Action: user_medias_for_instagram_cheapest
Action Input: {\"user_id\": \"113294420064920\"}"}]},
{"role": "function","content": [{"type": "text","text": "{\"error\": \"\", \"response\": \"{\\\"data\\\": [{\\\"media_id\\\": \\\"1234567890\\\", \\\"caption\\\": \\\"Just Do It\\\" ..."}]}]},
{"role": "assistant","content": [{"type": "text","text": "Thought: I have successfully retrieved the recent media posts ...
Action: Finish
Action Input: {\"return_type\": \"give_answer\", \"final_answer\": \"User information and recent media posts for 'nike' ...\"}"}]}] 
\end{lstlisting}
The full corpus, schema, and documentation are publicly released at: 
\url{https://huggingface.co/datasets/SriVatsa123/Gemma_Ready_For_PALADIN/viewer/default/train?views%5B%5D=train&row=0}.
\clearpage
\section{Recovery Dictionary}
\label{app:recover_dic}
At the core of PALADIN’s learning process is a recovery dictionary—a curated collection of over 50 execution-level failure types and corresponding recovery strategies. This dictionary was not arbitrarily constructed. Instead, it was built through an extensive review of real-world sources, including:
\begin{itemize}
    \item Developer forums (e.g., Stack Overflow, GitHub Issues)
    \item Toolchain documentation (e.g., LangChain, Zapier, API reference manuals)
    \item Industry engineering blogs detailing agent failure cases (e.g., Google’s Bard, OpenAI function calls)
    \item Academic surveys and benchmarks (e.g., ToolScan, ShieldA, Healer)
\end{itemize}
We structured our final recovery dictionary around ToolScan’s taxonomy of common agentic failures, enriched with patterns synthesized from forums, documentation, and engineering logs. Each error type in the dictionary is paired with recovery actions grounded in real debugging and fallback strategies reported by practitioners.

These error-response mappings were then transformed into example trajectories using the ToolBench format, providing PALADIN with richly annotated failure contexts and realistic correction paths. By rooting this dictionary in actual agent failure logs and developer strategies, we ensured the training data reflected authentic, actionable recovery behavior, not synthetic artifacts.

\section{ToolScan Error Types}
\label{app:error_t}
\textit{Tool Hallucination} → tool not found errors

\textit{Argument Hallucination} → missing or invalid tool input

\textit{Invalid Tool Invocation} → tool call failed

\textit{Partial Execution} → incomplete tool outputs requiring continuation

\textit{Output Hallucination} → tool output not valid

\textit{Invalid Intermediate Reasoning} → corrected plans and backtracking in multi-turn dialogues

\textit{Re-entrant Failures} → handled via explicit retry and fallback logic
\clearpage
\section{Python Dictionary}
\label{app:python_dic}
Here is part of our dictionary: 

\lstset{
  basicstyle=\ttfamily\small,
  keywordstyle=\color{blue}\bfseries,
  stringstyle=\color{teal},
  commentstyle=\color{gray},
  showstringspaces=false,
  breaklines=true,
  frame=single,
}

\begin{lstlisting}[language=Python, caption={Recovery Paths Dictionary}]
recovery_paths = {

  # Branch 1: Client Request Syntax and Format Errors (400, 422)
  "400_422": [
    {
      "from": "Assistant",
      "value": (
        "Thoughts: The 400 Bad Request and 422 Unprocessable Entity errors indicate client-side issues. 400 "
        "often means malformed syntax or invalid request formatting; 422 means the syntax is correct but the "
        "content is semantically invalid.\n\n"
        "Action: Start by examining the request URL and endpoint carefully. Check for typos, extra slashes, or "
        "illegal characters that violate URL standards. Use URL validators or linters. Then, validate your "
        "payload format; if JSON, ensure it is well-formed, properly bracketed, and matches the API schema. Check "
        "required fields and types.\n\n"
        "Verify that all required headers like Content-Type and Authorization are present and correct. For 422, "
        "focus on semantic validation such as correct value ranges, formats, and dependencies."
      )
    },
    {
      "from": "function",
      "value": "Validated URL and headers; payload verified with no syntax errors."
    },
    {
      "from": "Assistant",
      "value": (
        "Thoughts: If syntax is good, server-side validation may enforce strict formats or rules not obvious in docs.\n\n"
        "Action: Test with minimal payloads including only required fields; gradually add optional fields to isolate "
        "problematic data. Confirm API version matches docs, and check server logs if available."
      )
    },
    {
      "from": "function",
      "value": "Minimal payload testing performed; error persists."
    },
    {
      "from": "Assistant",
      "value": (
        "Thoughts: Persistent error suggests possible issues with query parameters or headers.\n\n"
        "Action: Verify all query parameters for correctness and allowed values. Confirm proper encoding (UTF-8) "
        "and escaping of special characters. Check multipart boundaries if uploading files."
      )
    },
    {
      "from": "function",
      "value": "Query parameters and encoding verified."
    },
    {
      "from": "Assistant",
      "value": (
        "Thoughts: If no client-side issue found, consider API bugs or server incompatibility.\n\n"
        "Action: Contact support with detailed logs; implement retry with exponential backoff."
      )
    },
    {
      "from": "function",
      "value": "Escalation logs prepared; retry logic implemented."
    }
  ],

  # Branch 2: Authentication and Authorization Errors (401, 403, 407)
  "401_403_407": [
    {
      "from": "Assistant",
      "value": (
        "Thoughts: 401 Unauthorized indicates missing or invalid credentials, 403 Forbidden means access is denied "
        "even if authenticated, and 407 Proxy Authentication Required means the proxy server requires authentication.\n\n"
        "Action: Check that valid API keys or tokens are sent in headers exactly as required by the service. Refresh "
        "tokens if expired. Confirm permissions on the user or API key to access requested resources.\n\n"
        "For 403, verify role-based access controls and resource-specific permissions. For 407, ensure proxy credentials "
        "are configured and included properly in requests."
      )
    },
    {
      "from": "function",
      "value": "Credentials and permissions checked; tokens refreshed where needed."
    },
    {
      "from": "Assistant",
      "value": (
        "Thoughts: Persistent 401/403 might be due to time skew causing token rejection or revoked credentials.\n\n"
        "Action: Sync system clock with NTP, regenerate credentials, and check for IP restrictions or rate limits "
        "that may block access."
      )
    },
    {
      "from": "function",
      "value": "System time synchronized; no IP blocks detected."
    },
    {
      "from": "Assistant",
      "value": (
        "Thoughts: If proxy authentication fails, check proxy server logs and credentials.\n\n"
        "Action: Update proxy config or switch to a proxy that supports needed authentication methods."
      )
    },
    {
      "from": "function",
      "value": "Proxy credentials updated; connection succeeded."
    }
  ],

}
\end{lstlisting}
The rest of our dictionary can be found in github at https://github.com/HexaA2/paladin. 

\section{Implementation Details}
\label{app:implementation}

\subsection{Model Instantiations}
\textsc{PALADIN} was fine-tuned on multiple open-weight backbones to test cross-architecture generality: Gemma-27B, Qwen-2.5-14B-Instruct, AM-Thinking V1, and LLaMA-3.1-8B-Instruct. All models were trained on the full 50K recovery-annotated corpus, enabling direct comparison of robustness transfer across scales and inductive biases. For LLaMA-3.1-8B, extended context via RoPE scaling ensured parity on long recovery traces relative to Gemma and Qwen, avoiding truncation artifacts.

\subsection{LoRA Configuration}
All runs adopted LoRA adapters with rank $16$, scaling $\alpha = 32$, and dropout $0.0$. Adapters were injected into attention projections (\texttt{q\_proj}, \texttt{k\_proj}, \texttt{v\_proj}, \texttt{o\_proj}) and MLP projections (\texttt{up\_proj}, \texttt{down\_proj}, \texttt{gate\_proj}), equipping models with recovery skills while preserving base competence.

\subsection{Optimizer and Scheduler}
Training employed paged AdamW (32-bit) with \texttt{bf16} precision and gradient checkpointing. We used a base learning rate of $2 \times 10^{-4}$, with standard AdamW defaults for warmup and weight decay, and a constant schedule over one epoch, appropriate for single-epoch SFT with LoRA.

\subsection{Batching, Context, and Epochs}
Experiments used a context length of $8192$ tokens. Training was performed with micro-batch size $1$ and gradient accumulation $8$ (effective batch size $8$), for a single epoch over 50K trajectories (80\% failure-rich, 20\% clean ``happy paths''). This composition balanced recovery supervision with baseline tool-use competence.

\subsection{Hardware and Runtime}
All fine-tuning ran on \texttt{h200sxm} GPUs. The combination of \texttt{bf16}, paged AdamW, and checkpointing enabled stable 8K-token SFT with LoRA on these accelerators. RoPE scaling extended LLaMA-3.1-8B’s effective context to $128$K tokens with modest additional memory cost, remaining tractable under \texttt{h200}-class footprints. End-to-end fine-tuning completed within a single epoch per backbone, with wall-clock time increasing monotonically with parameter count ($8$B $<$ $14$B $<$ $27$B).

\clearpage
\section{Dataset Construction Details}
\label{app:dataset}

\subsection{Failure Injection Procedure}
We began from ToolBench tasks and tool schemas, discarding original rollouts, and applied an automated trace parser to detect the first execution error. Each trajectory was truncated at that failure point to create a repair target. For every truncated trace, a controller supplied the task, tool schema, error signal, and dialogue context to a GPT-5 Teacher equipped with a recovery dictionary. The Teacher then generated multi-turn \texttt{Recovery:} segments consisting of retries, reformulations, fallbacks, or graceful termination, producing repaired trajectories:
\[
f_{\text{repair}}(T, A, C, E) \;\;\to\;\; C' ,
\]
while error-free traces were finalized via:
\[
f_{\text{finalize}}(T, A, C) \;\;\to\;\; C' .
\]
The simulator injected failures deterministically by specifying error type, manifestation (e.g., malformed output, silent failure), and turn index. A Python controller executed each scenario by providing tool documentation, applying the designated error, and simulating tool responses to the agent’s recovery actions.

\paragraph{Injected error examples.}
\begin{itemize}
    \item \textbf{Timeouts and 5xx/503:} Transient server failures triggering capped backoff-and-retry before graceful termination.
    \item \textbf{Malformed/invalid outputs:} Truncated JSON, schema violations, or null fields designed to elicit re-queries, lenient parsing, or tool switching.
    \item \textbf{Auth/permission errors (401/403/407):} Non-recoverable or credentials-refresh scenarios; repeated failures prompted terminate-with-explanation policies.
\end{itemize}

\subsection{Recovery Annotation Process}
Recovery supervision combined two sources: (i) a curated recovery dictionary aligned to ToolScan’s seven error classes, and (ii) GPT-guided rewriting conditioned on truncated traces and error signals. For each failure, the Teacher expanded dictionary-level strategies into situated multi-turn recoveries in ToolBench format (\texttt{Thought} $\to$ \texttt{Action} $\to$ \texttt{Action Input} $\to$ \texttt{Tool Output}), prefixing corrective steps with \texttt{Recovery:} tags and concluding with \texttt{Finish} plus either a user-facing answer or graceful-failure explanation. Clean ``happy path'' traces (about 20\%) were also audited and, when necessary, rewritten to ensure fully successful interactions, preserving base competence while keeping recovery central in the remaining 80\% of traces.

\subsection{Recovery Exemplar Distribution}
PALADIN’s retrieval bank contains over 55 recovery exemplars derived from the dictionary and aligned to ToolScan’s seven canonical error types (see (see Appendix~\ref{app:recover_dic})). These cover \emph{Tool Hallucination}, \emph{Argument Hallucination}, \emph{Invalid Tool Invocation}, \emph{Partial Execution}, \emph{Output Hallucination}, \emph{Invalid Intermediate Reasoning}, and \emph{Re-entrant Failures}, each paired with exemplar failures and recovery protocols. During execution, observed failures are matched to the closest exemplar via signature distance $d$, with the associated recovery action steering trajectories back to stability. This enables generalization across diverse failure surfaces.

\subsection{Data Split}
The final corpus comprises roughly 50K trajectories serialized in ToolBench format with explicit \texttt{Recovery:} tags, with an 80/20 composition of recovery-rich to clean traces. Training sequences were processed under a single-epoch CLM SFT regime. Evaluations were conducted in a sandboxed environment with deterministic error injection to ensure controlled, reproducible assessment using TSR, RR, CSR, and Efficiency metrics. For LLaMA-3.1-8B, RoPE scaling extended effective context to 128K tokens, preventing truncation of long recovery trajectories and ensuring parity across backbones under the same split design.

\clearpage
\section{PaladinEval Benchmark}
\label{app:paladineval}

PaladinEval is a deterministic failure-injection benchmark designed to evaluate recovery competence, honesty, and efficiency under the seven ToolScan error classes. It combines controlled simulators, taxonomic labeling, and standardized metrics (TSR, RR, CSR, ES) to enable apples-to-apples comparison across models and methods. Unlike ToolReflectEval, which targets single-call reflective corrections, PaladinEval emphasizes trajectory-level, multi-turn recovery in noisy execution settings. All reported scores are normalized so that higher is better across tables and plots.

\subsection{Benchmark Design}
PaladinEval instruments ToolBench-style tasks with a simulator that injects a single, labeled execution failure at a specified turn, then drives the episode to completion while logging recovery attempts, retries, tool switches, and termination decisions. Each episode includes the task specification, tool schema, truncated trace at failure, and an injected error drawn from the ToolScan taxonomy. Evaluation is conducted with fixed seeds and deterministic tool outputs to ensure reproducibility across backbones and runs.

\subsection{Tasks and Coverage}
The suite spans diverse tool-use tasks representative of training domains and failure surfaces. Episodes are constructed to cover all seven ToolScan error categories under uniform sampling rules, preventing skew toward any particular class. Results reported in the main paper compare PaladinEval against ToolReflectEval across multiple backbones, showing consistent gains in Recovery Rate, Task Success Rate, and Catastrophic Success Rate. These improvements come with expected efficiency trade-offs from retry-heavy strategies, but confirm sufficient breadth and balance across failure classes for comparative evaluation.

\subsection{Sampling Procedure}
For each task, the first tool failure is injected deterministically by specifying the error class, manifestation (e.g., 5xx timeout, malformed JSON), and turn index. The remainder of the episode is executed with fixed tool responses to recovery actions, minimizing variance. Sampling ensures per-class coverage across \emph{Tool Hallucination}, \emph{Argument Hallucination}, \emph{Invalid Invocation}, \emph{Partial Execution}, \emph{Output Hallucination}, \emph{Invalid Intermediate Reasoning}, and \emph{Re-entrant Failures}, enabling both per-class and macro-averaged reporting of RR, TSR, CSR, and ES.

\subsection{Adapting ToolReflectEval}
For comparability, ToolReflectEval was re-run under the same deterministic simulator and normalized metrics. While ToolReflectEval emphasizes critique-based improvements to single tool calls, PaladinEval stresses multi-turn recovery after explicit failures, capturing behaviors such as diagnosis, replanning, retries, tool switches, and graceful termination. Together, the two benchmarks provide complementary perspectives on robustness.

\subsection{Filtering and Deduplication}
Benchmark construction applies truncation at the first failure and removes trajectories with ambiguous or duplicate error signatures to avoid double-counting or conflating error classes. Episodes with inconsistent tool schemas or non-reproducible outputs are excluded to preserve determinism. Clean “happy-path” episodes are retained for competence checks but not scored as recoveries, ensuring evaluation focuses squarely on execution robustness.

\clearpage
\section{Additional Metrics and Ablations}
\label{app:metrics-ablations}

\subsection{Metric Variants and Formulas}
We evaluate \textsc{PALADIN} with four metrics:

\begin{align*}
\text{Task Success Rate (TSR)} &= \tfrac{\text{\# successful tasks}}{\text{\# total tasks}}, \\
\text{Recovery Rate (RR)} &= \tfrac{\text{\# failures recovered}}{\text{\# failures encountered}}, \\
\text{Catastrophe Success Rate (CSR)} &= 1 - \tfrac{\text{\# hallucinated successes}}{\text{\# total failures}}, \\
\text{Efficiency Score (ES)} &= \tfrac{1}{\text{average \# steps to complete task}}.
\end{align*}

RR, CSR, and ES are novel contributions that capture execution-level robustness beyond traditional task success. For diagnostic checks, we also experimented with call-level efficiency and normalized efficiency variants; these preserved model rankings and are omitted from main results for clarity.

\subsection{Ablation: Removing Inference-Time Retrieval}
\textsc{PALADIN} uses taxonomic retrieval over a curated bank of 55+ recovery exemplars aligned to ToolScan to map observed failures to prototypical recovery actions. Removing retrieval disables exemplar matching and forces purely end-to-end behavior. Across backbones, this sharply reduces robustness:

\begin{itemize}
    \item \textbf{Gemma-12B:} RR 89.7\% $\to$ 61.4\%, TSR 87.4\% $\to$ 57.3\%, CSR 82.6\% $\to$ 65.1\%.
    \item \textbf{Qwen-14B:} RR 94.7\% $\to$ 73.3\%, TSR 79.5\% $\to$ 66.3\%, CSR 94.6\% $\to$ 68.9\%.
    \item \textbf{LLaMA-8B:} RR 79.8\% $\to$ 48.6\%, TSR 78.7\% $\to$ 42.7\%, CSR 80.7\% $\to$ 57.4\%.
    \item \textbf{AM-Thinking V1:} RR 96.1\% $\to$ 81.2\%, TSR 81.2\% $\to$ 70.9\%, CSR 88.7\% $\to$ 73.3\%.
\end{itemize}

Drops of 20–30 points highlight that learned recovery patterns are substantially amplified by exemplar-guided retrieval at inference.

\subsection{Ablation: Training Data Composition}
\textsc{PALADIN} is trained on recovery-rich traces (80\%) plus clean ``happy-path'' traces (20\%) to preserve tool competence and avoid overfitting to failure-only dynamics. Training solely on injected failures (without Teacher-authored \texttt{Recovery:} continuations) degrades multi-turn behavior: agents overfit to ``retry-once'' heuristics, miss plan-shift and tool-switch transitions, and exhibit decreased CSR due to missing supervised end states. Quantitatively, RR and TSR fall relative to full recovery-annotated training, with larger ES penalties from inefficient repeated retries. This confirms that explicit recovery annotations are essential for learning stable, compositional recovery behaviors.

\subsection{Robustness: Zero-Shot Transfer to Unseen Tools}
To evaluate generalization, we tested on held-out APIs under the same simulator and error taxonomy. Zero-shot transfer preserves a large fraction of recovery performance, with smaller absolute drops in CSR than RR/TSR. This suggests \textsc{PALADIN} preserves honesty under uncertainty even when recovery is incomplete. Failure analyses show strong transfer of schema-mismatch handling and malformed-output repair, while tool-specific authentication and pagination account for most residual errors. Retrieval mitigates these by guiding toward nearest exemplars, even when the exact tools are unseen.

\subsection{Takeaways}
\begin{itemize}
    \item Inference-time retrieval is a primary driver of robustness, complementing training-time exposure to diverse failures.
    \item Recovery-annotated trajectories are critical; failure-only training under-specifies chained recovery, degrading end-to-end completion and safety.
    \item Zero-shot transfer demonstrates that execution-level recovery behaviors generalize, with retrieval providing a safety net for novel APIs.
\end{itemize}

\section{Expanded Figures}
\label{app:expanded}
\begin{figure}[h]
\begin{center}
\includegraphics[width=1\linewidth]{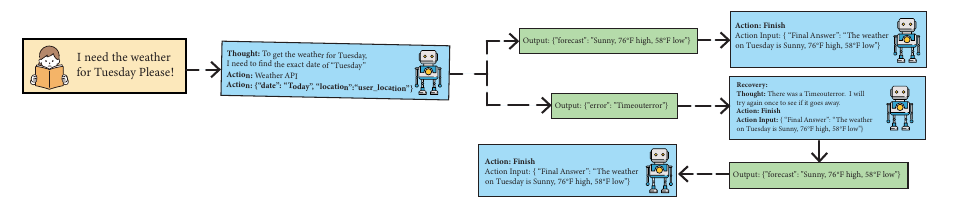}
\end{center}
\caption{PALADIN's Thought Process}
\label{fig:failure_injection}
\end{figure}

\begin{figure}[!htbp]
    \centering
    \begin{minipage}{0.48\linewidth}
        \centering
        \includegraphics[width=\linewidth]{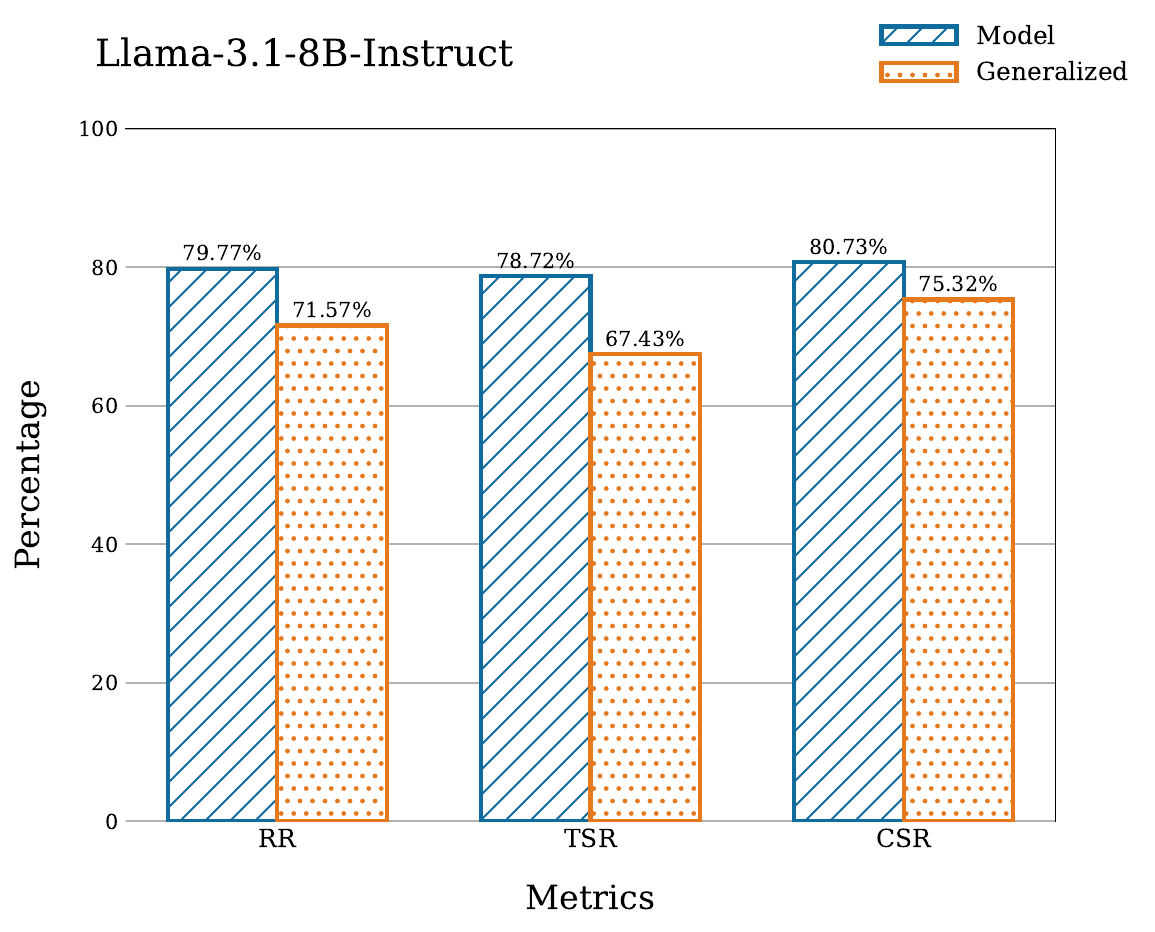}
        \caption*{(a) LLaMA-3.1-8B}
        \label{fig:Abalation-llama}
    \end{minipage}\hfill
    \begin{minipage}{0.48\linewidth}
        \centering
        \includegraphics[width=\linewidth]{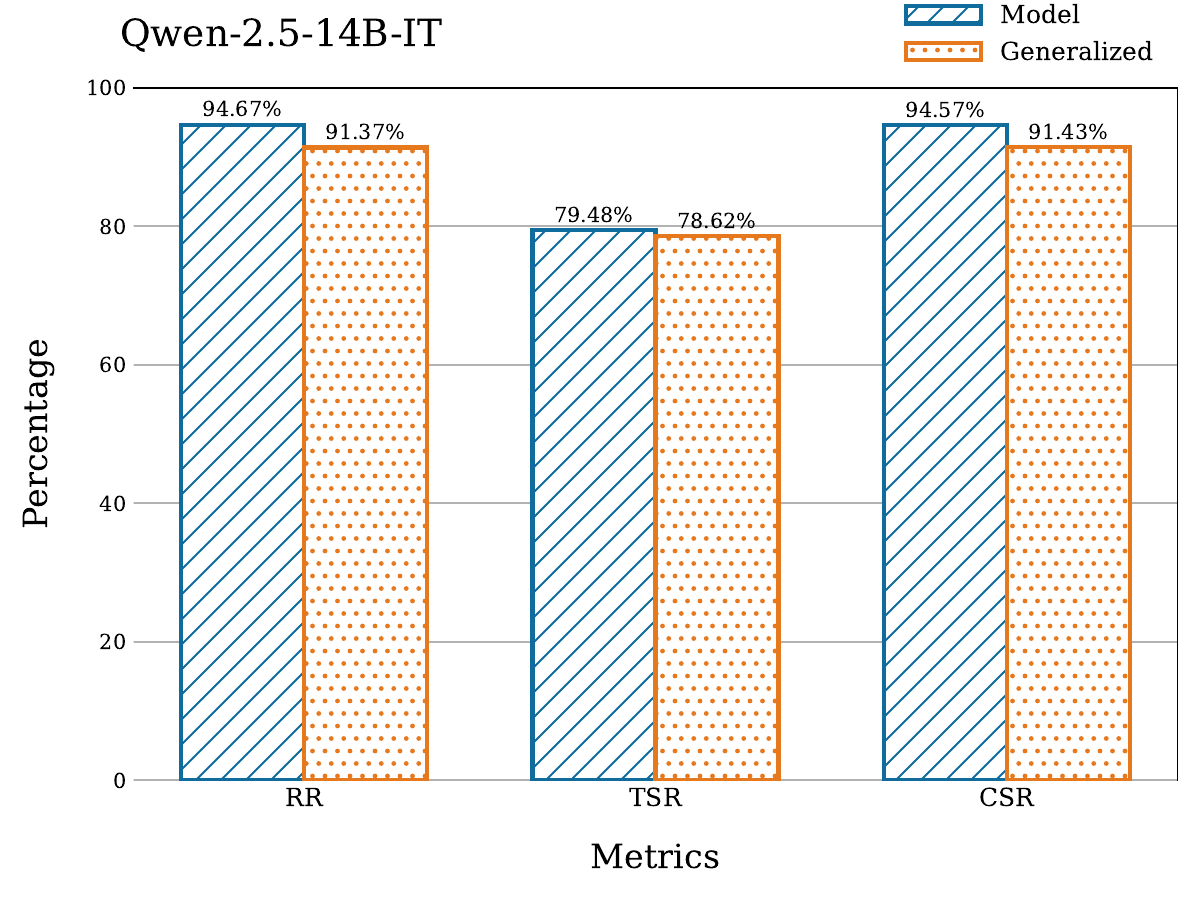}
        \caption*{(b) Qwen-2.5-14B}
        \label{fig:Abalation-qwen}
    \end{minipage}
    
    \vspace{0.3cm}
    
    \begin{minipage}{0.48\linewidth}
        \centering
        \includegraphics[width=\linewidth]{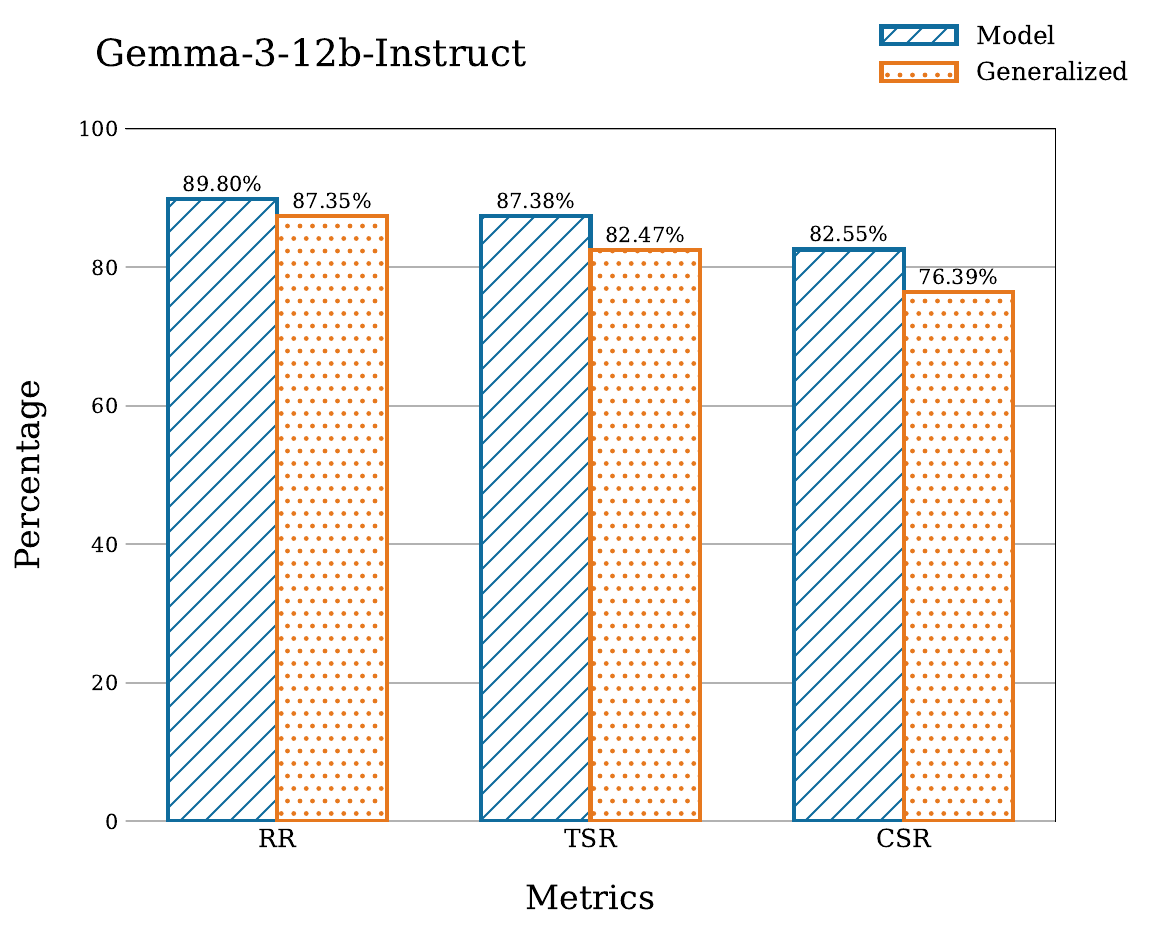}
        \caption*{(c) Gemma-3-12B}
        \label{fig:Abalation-gemma}
    \end{minipage}\hfill
    \begin{minipage}{0.48\linewidth}
        \centering
        \includegraphics[width=\linewidth]{iclr2026/AMThinkingGeneral.pdf}
        \caption*{(d) AM-Thinking-V1}
        \label{fig:Abalation-am}
    \end{minipage}
    
    \caption{PALADIN's robustness when facing unseen error types across different model backbones.}
    \label{fig:generalization}
\end{figure}

\begin{figure}[!htbp]
    \centering
    \begin{minipage}{0.48\linewidth}
        \centering
        \includegraphics[width=\linewidth]{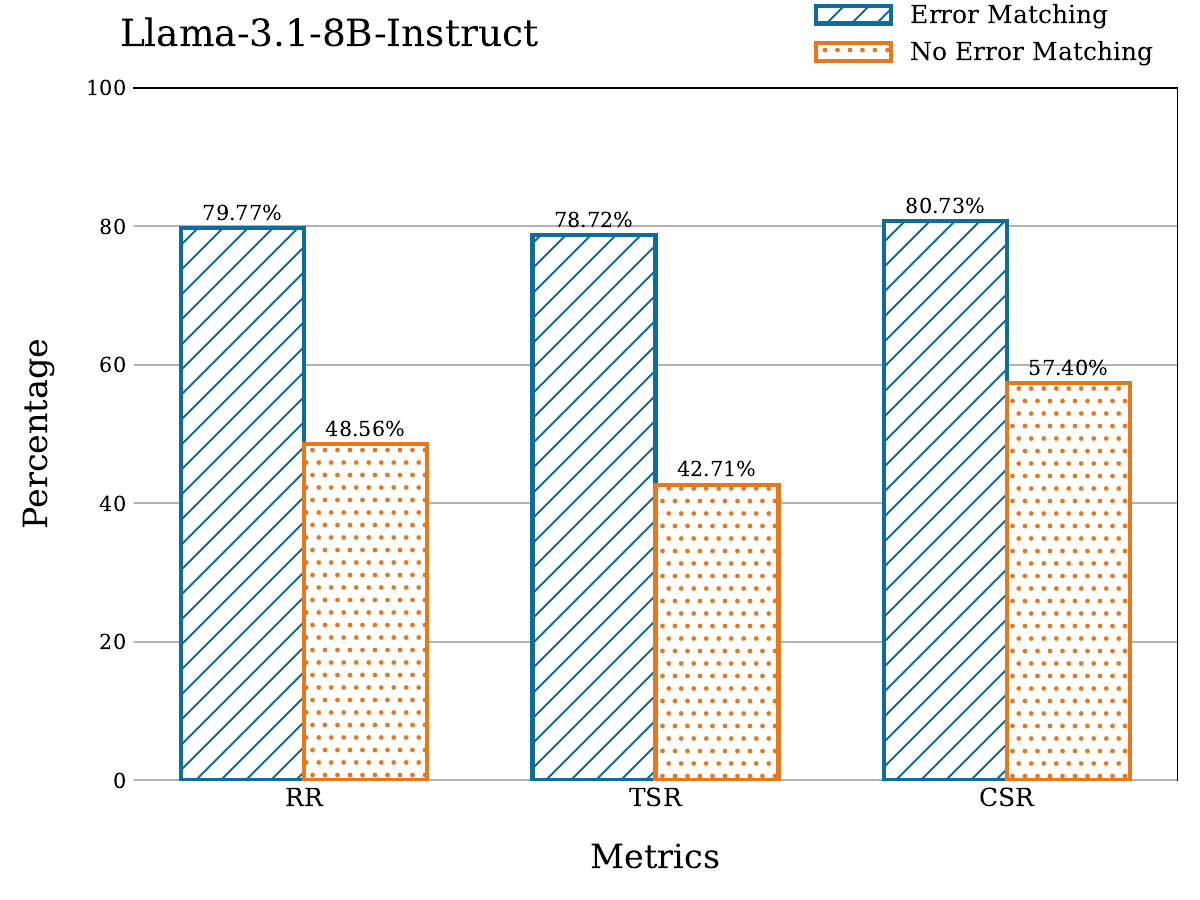}
        \caption*{(a) LLaMA-3.1-8B}
        \label{fig:Abalation-llama}
    \end{minipage}\hfill
    \begin{minipage}{0.48\linewidth}
        \centering
        \includegraphics[width=\linewidth]{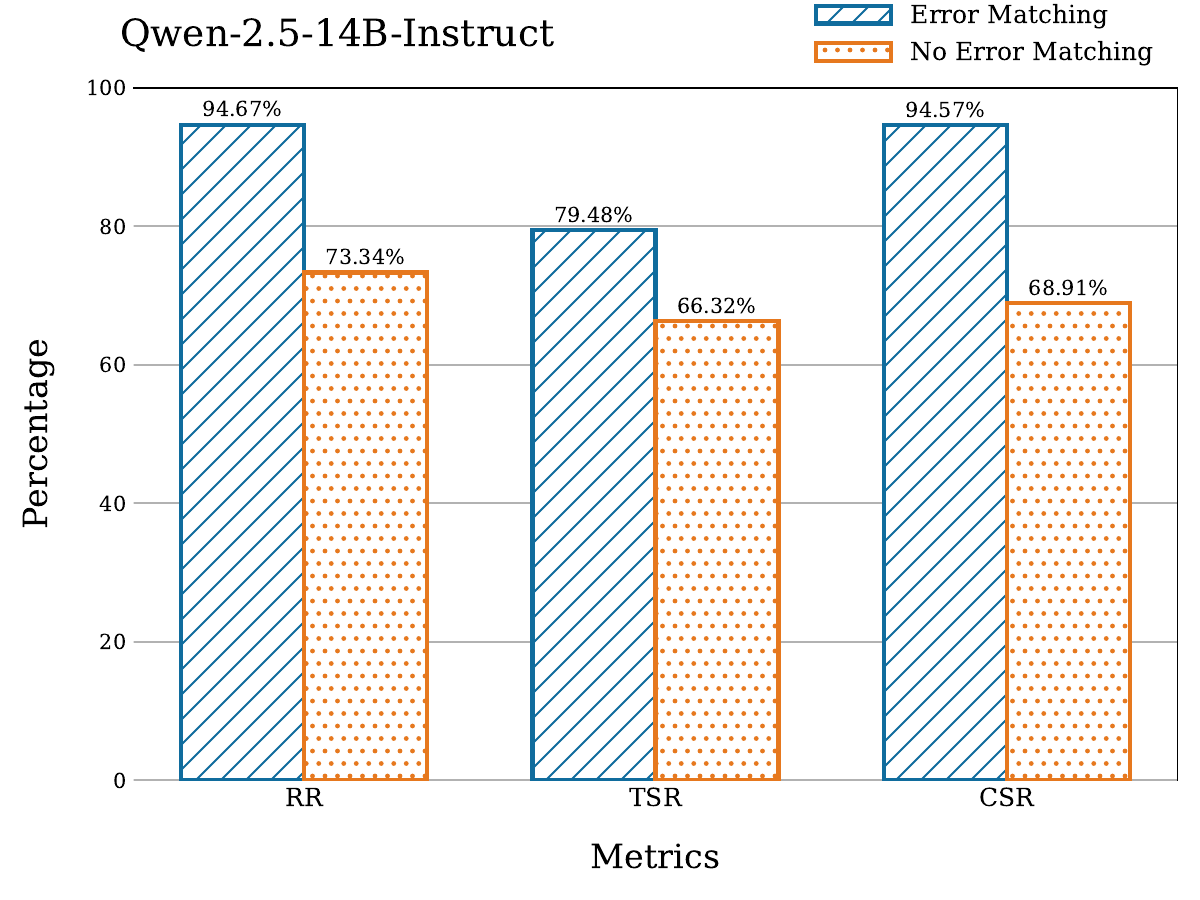}
        \caption*{(b) Qwen-2.5-14B}
        \label{fig:Abalation-qwen}
    \end{minipage}
    
    \vspace{0.3cm}
    
    \begin{minipage}{0.48\linewidth}
        \centering
        \includegraphics[width=\linewidth]{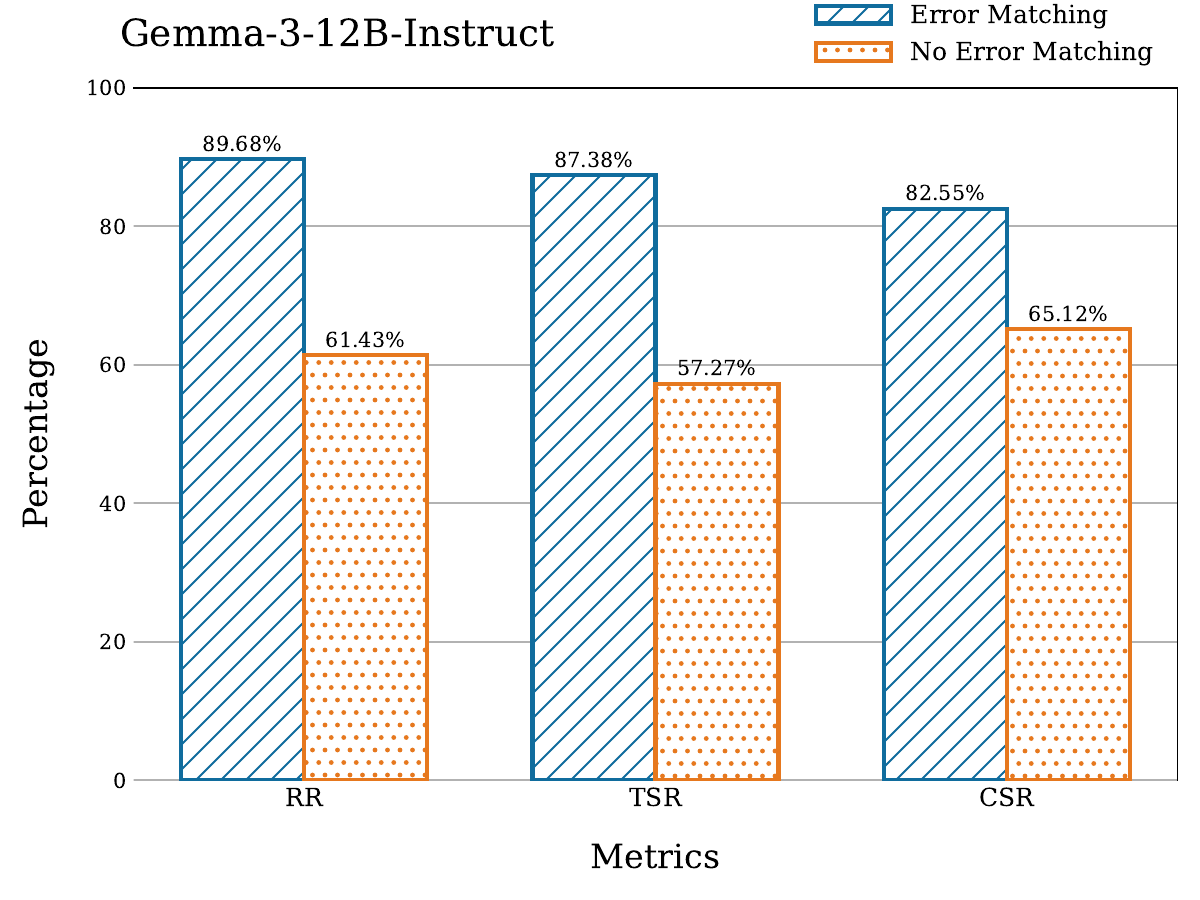}
        \caption*{(c) Gemma-3-12B}
        \label{fig:Abalation-gemma}
    \end{minipage}\hfill
    \begin{minipage}{0.48\linewidth}
        \centering
        \includegraphics[width=\linewidth]{iclr2026/AMThinkingAbal.pdf}
        \caption*{(d) AM-Thinking-V1}
        \label{fig:Abalation-am}
    \end{minipage}
    
    \caption{PALADIN's abalation when facing unseen error types across different model backbones.}
    \label{fig:Abalation-amthinking}
\end{figure}

\end{document}